\documentclass[10pt,journal,compsoc]{IEEEtran}

\usepackage{cite}
\usepackage{graphicx}
\usepackage{amsmath,amssymb,amsfonts}
\usepackage{algpseudocode} 
\usepackage{algorithm}
\usepackage{textcomp}
\usepackage{xcolor}
\usepackage{multirow}
\usepackage{booktabs}
\usepackage{url}
\usepackage{hyperref}
\hypersetup{colorlinks=true,linkcolor=blue,citecolor=blue,urlcolor=blue}

\usepackage{stfloats}
\usepackage{url}
\usepackage{verbatim}
\usepackage{threeparttable}
\usepackage{tabularx}
\usepackage{longtable}
\usepackage{graphicx}
\usepackage{multirow}
\usepackage{pifont}
\usepackage{titlesec}
\usepackage{booktabs}
\usepackage{algpseudocode}
\usepackage{graphicx}
\usepackage{subcaption}

\usepackage[most]{tcolorbox}
\usepackage{tikz}
\usepackage{listings}
\usepackage{enumitem}
\usepackage{xcolor}
\usepackage{ragged2e}

\definecolor{boxblue}{RGB}{222,239,255}
\definecolor{boxblueframe}{RGB}{60,120,200}
\definecolor{boxred}{RGB}{255,232,232}
\definecolor{boxredframe}{RGB}{200,60,60}
\definecolor{boxyellow}{RGB}{255,248,225}
\definecolor{boxyellowframe}{RGB}{210,170,60}

\tcbset{
  promptbox/.style={
    enhanced,
    sharp corners,
    boxrule=0.6pt,
    left=6pt,right=6pt,top=6pt,bottom=6pt,
    coltitle=black,
    fonttitle=\bfseries,
  }
}

\setlist[itemize]{leftmargin=1.2em,itemsep=2pt,topsep=2pt}

\lstdefinelanguage{json}{
  basicstyle=\ttfamily\small,
  showstringspaces=false,
  breaklines=true,
  literate=
   *{0}{{{\color{black}0}}}{1}
    {1}{{{\color{black}1}}}{1}
    {2}{{{\color{black}2}}}{1}
    {3}{{{\color{black}3}}}{1}
    {4}{{{\color{black}4}}}{1}
    {5}{{{\color{black}5}}}{1}
    {6}{{{\color{black}6}}}{1}
    {7}{{{\color{black}7}}}{1}
    {8}{{{\color{black}8}}}{1}
    {9}{{{\color{black}9}}}{1}
    {:}{{{\color{black}{:}}}}{1}
    {,}{{{\color{black}{,}}}}{1}
    {"}{{{\color{black}{"}}}}{1}
}

\begin{document}

\title{Safe and Economical UAV Trajectory Planning in Low-Altitude Airspace: A Hybrid DRL-LLM Algorithm with Compliance Awareness}

\author{Yanwei~Gong, Junchao~Fan, Ruichen~Zhang, Dusit~Niyato,~\IEEEmembership{Fellow,~IEEE}, Yingying~Yao, and Xiaolin~Chang
\IEEEcompsocitemizethanks{
\IEEEcompsocthanksitem Yanwei Gong, Junchao Fan, Yingying Yao, and Xiaolin Chang are with the Beijing Key Laboratory of Security and Privacy in Intelligent Transportation, Beijing Jiaotong University, P.R. China.
\protect\\ E-mail: \{22110136, 23111144, yyyao, xlchang\}@bjtu.edu.cn
\IEEEcompsocthanksitem Ruichen Zhang and Dusit Niyato are with the School of Computer Science and Engineering, Nanyang Technological University, Singapore.
\protect\\ E-mail: \{ruichen.zhang, dniyto\}@ntu.edu.sg
}}

\IEEEtitleabstractindextext{
\begin{abstract}
\justifying
The rapid growth of the low-altitude economy has driven the widespread adoption of unmanned aerial vehicles (UAVs). This growing deployment presents new challenges for UAV trajectory planning in complex urban environments. However, existing studies often overlook key factors, such as urban airspace constraints and economic efficiency, which are essential in low-altitude economy contexts. Deep reinforcement learning (DRL) is regarded as a promising solution to these issues, while its practical adoption remains limited by low learning efficiency. To overcome this limitation, we propose a novel UAV trajectory planning algorithm that integrates DRL with the large language model (LLM) reasoning to enable safe, compliant, and economically viable trajectory planning. Specifically, we model the trajectory planning task as a partially observable Markov decision process, explicitly incorporating obstacle avoidance, regulation awareness, and energy constraints. We design a hybrid optimization algorithm based on the soft actor-critic algorithm and LLM reasoning to enable adaptive decision-making in uncertain and dynamic environments. Experimental results demonstrate that our algorithm achieves the best overall performance, with the highest data collection rate (99.50\%), almost zero collision avoidance rate and regulation violation rate, perfect successful landing rate (100\%), and the lowest energy consumption rate (76.95\%). These results validate the effectiveness of our algorithm in addressing UAV trajectory planning key challenges under constraints of the low-altitude economy networking.
\end{abstract}

}

\maketitle


\section{Introduction}
With the rapid advancement of the low-altitude economy, low-altitude airspace has emerged as a critical strategic resource to drive urban intelligence, industrial upgrading, and digital economic transformation~\cite{15,16}. This evolving domain integrates diverse sectors such as general aviation and urban air mobility, wherein unmanned aerial vehicles (UAVs) play an increasingly vital role in applications including logistics, agricultural monitoring, infrastructure inspection, and environmental protection~\cite{2}. However, the high-density and regulation-intensive nature of low-altitude airspace also poses unique challenges for intelligent trajectory planning and safe autonomous operation.

In this context, UAV-based data acquisition represents a fundamental technology enabling high-precision, multi-modal environmental sensing~\cite{18}. The UAV data acquisition process typically involves task assignment, flight trajectory planning, sensor data collection, and data transmission and processing. Effective trajectory planning is crucial to maximize area coverage and data quality, thereby supporting downstream applications with accurate and timely information~\cite{17}. It also serves as a key enabler for intelligent and autonomous decision-making in dense and dynamic airspaces.

Despite its potential, trajectory planning for UAV data acquisition in low-altitude airspace faces the following important challenges:

\textbf{Challenge 1: Robust Obstacle Avoidance in Urban Airspaces.} Due to the existence of airspace reuse in the low-altitude economy~\cite{36}, operating in low-altitude urban environments exposes the data collection UAVs (DCU) not only to static obstacles (e.g., buildings) but also to dynamic obstacles (e.g., other UAVs). These obstacles are often partially observable and highly uncertain. Moreover, the DCU must also avoid restricted no-fly zones (NFZs) over sensitive areas. Ensuring robust obstacle avoidance under such conditions remains a core challenge. Ensuring robust obstacle avoidance under such conditions remains a core challenge. Traditional optimization methods typically lack the adaptability to handle such stochastic environments, and standard deep reinforcement learning (DRL) struggles to accumulate sufficient successful trajectories, leading to low learning efficiency~\cite{5}.

\textbf{Challenge 2: Intelligent Planning for the Economy with Compliance.} In the low-altitude economy, the DCU must not only maximize data acquisition but also minimize energy consumption to ensure cost-effectiveness. Moreover, the DCU must adhere to evolving, location-specific regulations where the DCU is to reduce speed to minimize noise~\cite{6}. These decisions introduce additional requirements on mission efficiency and regulatory compliance. Traditional methods struggle with multi-objective adaptation under uncertainty, while conventional DRL approaches often fail to incorporate domain knowledge needed to make economically viable decisions.

\textbf{Challenge 3: Real-Time Decision-Making Under Uncertainty.} Trajectory planning in real-world scenarios requires real-time decision making across multiple dimensions under dynamic and uncertain conditions, including obstacle avoidance, route selection, and data collection. Traditional optimization methods are often computationally intensive and lack responsiveness, whereas standard DRL policies may require extensive exploration due to task complexity.

Most existing methods address only isolated objectives or assume simplified environments, which prevents them from achieving a unified balance among safety, compliance, and economy. To address these limitations, this paper proposes a novel hybrid DRL and large language model (LLM) algorithm for safe and economical DCU trajectory planning in low-altitude airspace with compliance-awareness. The motivation is that, while the LLM offers strong reasoning abilities and can infer safe actions based on high-level semantic input, it lacks the ability to learn optimal control policies through interaction with dynamic environments~\cite{33}. Conversely, DRL excels at learning adaptive control policies via trial and error, but struggles to incorporate abstract constraints such as regulatory rules~\cite{34}. Therefore, we introduce an LLM to enhance the DCU’s decision-making capability in complex, uncertain scenarios where rule compliance, contextual awareness, and obstacle interpretation are critical. The main contributions of this paper are summarized as follows:
\begin{itemize}
  \item \textbf{Unified modeling of low-altitude UAV trajectory planning.} 
  We formulate the DCU trajectory planning task for low-altitude
 data acquisition as a partially observable Markov decision process (POMDP), which jointly captures constraints including both static and dynamic obstacle avoidance mentioned in \textbf{Challenge 1}, regulation awareness, and energy efficiency mentioned in \textbf{Challenge 2}, under partial observability.

  \item \textbf{A hybrid SAC and LLM-based trajectory planning algorithm.} We develop a novel optimization algorithm that combines the soft actor-critic (SAC) algorithm with an LLM. By combining the adaptive decision-making capabilities of SAC with the contextual reasoning strength of the LLM, the proposed algorithm enables the DCU to perform trajectory planning in real time, with robust obstacle avoidance, regulatory compliance, and economical data collection in uncertainty, thus addressing \textbf{Challenge 3}.

  \item \textbf{Comprehensive experimental validation and performance improvement.} 
  Extensive experiments verify that the proposed algorithm consistently outperforms state-of-the-art baselines across all key metrics. 
  Compared with the best-performing baseline, our method improves the data collection rate by 2.9\%, reduces the collision rate and regulation violation rate to nearly 0\%, achieves a 100\% successful landing rate, and lowers the energy consumption rate by 1.9\%. 
  These results confirm the algorithm’s effectiveness in addressing the \textbf{Challenges 1-3}.
\end{itemize}

The remainder of this paper is organized as follows. Section 2 reviews the related work. Section 3 describes the overall system architecture, while Section 4 presents the formal problem definition. Section 5 introduces the POMDP-based modeling approach and elaborates on the proposed algorithm. Section 6 reports the experimental results and performance evaluations. Finally, Section 7 concludes the paper and outlines future research directions.

\section{Related Work}
In this section, we review existing UAV trajectory planning methods from three perspectives, including traditional optimization algorithms, DRL-based approaches, and LLM-enhanced methods. 
Table~\ref{table1} summarizes representative studies and their characteristics.

\subsection{UAV Trajectory Planning with Traditional Algorithms}
Early studies mainly adopted optimization-based algorithms to jointly optimize UAV trajectory, communication, and energy efficiency. Qin et al.~\cite{7} used Dinkelbach’s method and block coordinate descent (BCD) for joint trajectory and resource optimization in reconfigurable intelligent surface-assisted UAV mobile edge computing (MEC) systems. Pan et al.~\cite{8} combined an improved non-dominated sorting genetic algorithm II (NSGA-II) with a customized particle swarm optimization (PSO) variant for multi-objective power and trajectory planning. Pervez et al.~\cite{9} formulated a multi-UAV MEC optimization using game theory and successive convex approximation (SCA). Zhang et al.~\cite{10} applied differential evolution to balance UAV deployment and flight planning for Internet of Things (IoT) data collection. Heo et al.~\cite{11} incorporated NFZ constraints using quadratic transform and SCA. Other works~\cite{12,13,14} explored swarm intelligence and distributed optimization for beamforming and energy efficiency.

Traditional optimization algorithms are effective in structured or static environments but lack adaptability to dynamic and uncertain low-altitude airspace conditions, and they cannot efficiently handle the stochastic nature of real-world obstacle distributions (\textbf{Challenge~1}).

\subsection{UAV Trajectory Planning with DRL}
To improve adaptability, many studies employed DRL for UAV trajectory control and resource management. Silvirianti et al.~\cite{19} proposed a layerwise quantum DRL method for joint trajectory and power optimization. Chen et al.~\cite{20} applied DRL to solve a multi-stage mixed-integer problem for UAV-assisted MEC. Ning et al.~\cite{21} introduced distributed DRL with game theory for collaborative control. Song et al.~\cite{22} formulated a multi-objective Markov decision process (MDP) with a trace-based experience replay mechanism. Ding et al.~\cite{23} developed a multi-agent advantage actor-critic (A2C)–deep deterministic policy gradient (DDPG) scheme with attention-based inference. Wang et al.~\cite{24} incorporated safety constraints into DRL under a constrained MDP framework. He et al.~\cite{25} and Liu et al.~\cite{26} extended multi-agent DRL to cooperative trajectory planning. Ning et al.~\cite{27} adopted constrained DRL for UAV mobility and connection optimization in IoT networks.

DRL-based methods can adapt to partially observable and dynamic environments, but they often require extensive exploration to incorporate domain-specific rules such as regulatory or economic constraints (\textbf{Challenge~2} and \textbf{Challenge~3}).

\subsection{UAV Trajectory Planning with LLM}
Recent research has introduced LLMs to enhance UAV autonomy by integrating reasoning and natural language understanding. 
Zhong et al.~\cite{28} combined lightweight object detection with an LLM for vision-based UAV planning in dynamic scenes. 
Phadke et al.~\cite{29} explored LLM-driven UAV control via natural language interfaces for improved mission coordination and safety.
Xiao et al.~\cite{30} proposed a foundation model-guided trajectory framework integrating LLMs and vision-language models (VLMs) for semantic reasoning and perception.
Samma et al.~\cite{31} utilized an LLM-based vision system to plan UAV trajectories in GPS-denied indoor environments.
Cai et al.~\cite{32} designed an LLM–model predictive control (MPC) hybrid landing system for enhanced safety in unstructured scenarios.

LLM-enhanced approaches offer semantic reasoning and interpretability but face challenges in real-time decision-making for UAVs, including excessive resource consumption and limited real-time performance.

\subsection{Discussion}
Existing UAV trajectory planning methods contribute valuable insights but exhibit limitations when applied to low-altitude data collection missions. 
Traditional algorithms~\cite{7}-\cite{14} cannot adapt to dynamic and uncertain obstacle patterns in real urban airspaces. 
DRL-based methods~\cite{19}-\cite{27} improve adaptability but overlook regulatory compliance and economic trade-offs. 
LLM-based frameworks~\cite{28}-\cite{32} offer strong reasoning but lack real-time adaptability.
Consequently, none of these approaches can simultaneously ensure robust obstacle avoidance, compliance with evolving regulations, and energy-efficient trajectory design in low-altitude environments. 
To overcome these issues, this paper proposes a hybrid DRL–LLM algorithm that integrates adaptive control and semantic reasoning to achieve safe, compliant, and economical UAV trajectory planning.

\begin{table*}
\begin{center}
\begin{threeparttable}
\caption{Comparison of Related Works on UAV Trajectory Planning}
\label{table1}
\begin{tabular}{| c | r | c | c | c |}
\hline
 & \multicolumn{1}{c|}{Ref.} & Main Method / Algorithm & Obstacle Avoidance & Compliance \\
\hline
\multirow{8}{*}{\rotatebox{90}{Traditional}} 
& Qin et al. 2023~\cite{7}      & Dinkelbach’s method, BCD       & \ding{55} & \ding{55} \\
\cline{2-5}
& Pan et al. 2024~\cite{8}       & NSGA-II, PSO variant           & Static only & \ding{55} \\
\cline{2-5}
& Pervez et al. 2024~\cite{9}    & Game theory, SCA               & \ding{55} & \ding{55} \\
\cline{2-5}
& Zhang et al. 2024~\cite{10}    & Evolutionary algorithm         & \ding{55} & \ding{55} \\
\cline{2-5}
& Heo et al. 2024~\cite{11}      & Quadratic transform, SCA       & \ding{55} & \ding{55} \\
\cline{2-5}
& Li et al. 2024~\cite{12}       & Swarm intelligence-based algorithm                            & \ding{55} & \ding{55} \\
\cline{2-5}
& Fu et al. 2025~\cite{13}       & Distributed approximate
 Newton, BCD                      & \ding{55} & \ding{55} \\
\cline{2-5}
& Wang et al. 2025~\cite{14}     & Ant colony optimization                           & Static only & \ding{55} \\
\hline
\multirow{9}{*}{\rotatebox{90}{DRL}} 
& Silvirianti et al. 2024~\cite{19} & Layerwise quantum-DRL       & \ding{55} & \ding{55} \\
\cline{2-5}
& Chen et al. 2024~\cite{20}     & DRL                            & \ding{55} & \ding{55} \\
\cline{2-5}
& Ning et al. 2024~\cite{21}     & DRL, game theory               & \ding{55} & \ding{55} \\
\cline{2-5}
& Song et al. 2024~\cite{22}     & Multi-objective RL             & \ding{55} & \ding{55} \\
\cline{2-5}
& Ding et al. 2024~\cite{23}     & DDPG, DAI                      & \ding{55} & \ding{55} \\
\cline{2-5}
& Wang et al. 2024~\cite{24}     & Safe DRL                       & \ding{55} & \ding{55} \\
\cline{2-5}
& He et al. 2024~\cite{25}       & DRL                            & \ding{55} & \ding{55} \\
\cline{2-5}
& Liu et al. 2025~\cite{26}      & Multi-agent RL, SCA            & \ding{55} & \ding{55} \\
\cline{2-5}
& Ning et al. 2025~\cite{27}     & Constrained DRL                & Static only & \ding{55} \\
\hline
\multirow{5}{*}{\rotatebox{90}{LLM}} 
& Zhong et al. 2024~\cite{28}    & Apply LLM directly             & Dynamic only & \ding{55} \\
\cline{2-5}
& Phadke et al. 2024~\cite{29}   & Apply LLM directly             & Static only & \ding{55} \\
\cline{2-5}
& Xiao et al. 2025~\cite{30}     & LLM, VLM                       & Static only & \ding{55} \\
\cline{2-5}
& Samma et al. 2025~\cite{31}    & Fine-tuned LLM                 & Both dynamic and static & \ding{55} \\
\cline{2-5}
& Cai et al. 2025~\cite{32}      & LLM, MPC                       & Both dynamic and static & \ding{55} \\
\hline
\end{tabular}
\begin{tablenotes}
\footnotesize
\item Note: \ding{55} indicates not considered.
\end{tablenotes}
\end{threeparttable}
\end{center}
\end{table*}

\begin{figure}[!t]
\centering
\includegraphics[width=3.5in]{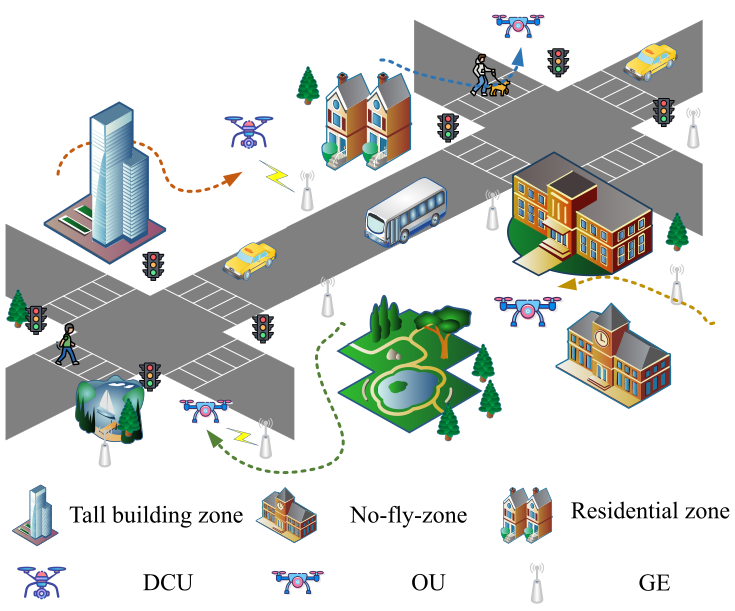}
\caption{System model of low-altitude UAV trajectory planning. The DCU collects data from GE while avoiding static obstacles (BZs, NFZs, RZs) and dynamic obstacles (OUs) under energy and compliance constraints.}
\label{fig1}
\end{figure}

\begin{table}[!t]
\renewcommand{\arraystretch}{1.2}
\caption{Notations and Descriptions}
\label{table2}
\centering
\begin{tabular}{p{3.2cm} p{4.8cm}}
\hline
\textbf{Notation} & \textbf{Description} \\
\hline
$\mathbb{A}=(X,Y,Z)$ & Area where the DCU conducts data collection, with length $X$, width $Y$, and height $Z$ \\
${B}_{\mathrm{GE}_{i}}(t)$ & Effective rate between the DCU and GE$_i$ in the time slot $t$ \\
$d_{t,\mathrm{XX}}$ & Distance between the DCU and XX at the beginning of time slot $t$ (XX = $\mathbb{A}$, $\mathbb{LA}$, OU, BZ, or NFZ) \\
$d_{\mathrm{min},\mathrm{XX}}$ & Minimum distance between the DCU and XX (XX = $\mathbb{A}$, OU, BZ, or NFZ) \\
$d_{\mathrm{safe},\mathrm{XX}}$ & Safe distance between the DCU and XX (XX = $\mathbb{A}$, OU, BZ, or NFZ) \\
$D{{V}_{G{{E}_{i}},t}}$ & Data volume of the $i$-th GE at the beginning of time slot $t$ \\
$E(t)$ & Remaining energy at the beginning of time slot $t$ \\
${{E}_{\text{limit}}}$ & Maximum energy that the DCU can consume \\
${{E}_{\mathrm{total}}}$ & Total energy the DCU has \\
$e(t)$ & Energy consumed in the time slot $t$ of the DCU \\
$e_{\mathrm{fly}}(t)$ & Flight energy consumption in the time slot $t$ of the DCU \\
$e_{\mathrm{hover}}(t)$ & Hover energy consumption in the time slot $t$ of the DCU \\
$\mathrm{EL}=({{x}_{\mathrm{EL}}},{{y}_{\mathrm{EL}}},{{z}_{\mathrm{EL}}})$ & Ending location of the DCU \\
$H$ & Fixed flight altitude of the DCU and OUs \\
$\mathbb{L}\mathbb{A}=({{p}_{\mathbb{L}\mathbb{A}}},{{l}_{\mathbb{L}\mathbb{A}}},{{w}_{\mathbb{L}\mathbb{A}}})$ & Landing area location of the DCU \\
${{L}_{{\mathrm{GE}}_{i}}}=({{x}_{{\mathrm{GE}}_{i}}},{{y}_{{\mathrm{GE}}_{i}}},0)$ & Location of the $i$-th GE \\
${{L}_{\mathrm{XX},t}}=({{x}_{\mathrm{XX},t}},{{y}_{\mathrm{XX},t}},{{z}_{\mathrm{XX},t}})$ & Location of XX at the beginning of time slot $t$ (XX = DCU or OU) \\
${{N}_{XX}}$ & Number of XX (XX = GE, OU, NFZ, BZ, or RZ) \\
$PR$ & Perception radius of the DCU \\
$\mathrm{SL}=({{x}_{\mathrm{SL}}},{{y}_{\mathrm{SL}}},{{z}_{\mathrm{SL}}})$ & Starting location of the DCU \\
${\mathrm{SNR}}_{\mathrm{GE}_i}(t)$ & The signal-to-noise ratio at the beginning of time slot $t$ of $\mathrm{GE}_i$ \\
$\mathbb{T}\mathbb{A}=({{p}_{\mathbb{T}\mathbb{A}}},{{l}_{\mathbb{T}\mathbb{A}}},{{w}_{\mathbb{T}\mathbb{A}}})$ & Take-off area location with lower-left coordinate point, length ${{l}_{\mathbb{T}\mathbb{A}}}$, and width ${{w}_{\mathbb{T}\mathbb{A}}}$ \\
$T$ & Total number of time slots \\
$TD_{{\mathrm{GE}}_{i}}(t)$ & Data transmitted between the DCU and GE$_i$ in the time slot $t$ \\
$T{{P}_{{\mathrm{GE}}_{i}}}$ & Transmit power of the $i$-th GE \\
${{t}_{\mathrm{fly}}}$ & Time used for flight in the time slot $t$ of the DCU \\
${{t}_{\mathrm{hover}}}$ & Time used for hovering in the time slot $t$ of the DCU \\
${v}_{\mathrm{DCU}}(t)=({{v}_{x}}(t),{v}_{y}(t),{{v}_{z}}(t))$ & Flight speed of the DCU at the beginning of time slot $t$ \\
${{v}_{\max}}$ & Maximum flight speed of the DCU and OUs \\
${{v}_{\text{limit}}}$ & Speed limit in ${{\mathbb{R}}_{\mathrm{RZ}}}$ for the DCU and OUs \\
$\wedge$ & Logical AND operator \\
$\odot$ & Hadamard (element-wise) product \\
$\left\| CP1-CP2/A \right\|$ & Distance between two coordinate points or to area $A$ \\
${{\mathbb{R}}_{\mathrm{XX}_{x}}}=({{p}_{\mathrm{XX}_{x}}},{{l}_{\mathrm{XX}_{x}}},{{w}_{\mathrm{XX}_{x}}})$ & Location of the $x$-th XX in $\mathbb{A}$ (XX = NFZ, BZ, or RZ) \\
$\Delta t$ & Fixed time slot duration \\
$\tau$ & The threshold of ${\mathrm{SNR}}_{\mathrm{GE}_i}(t)$ \\    
\hline
\end{tabular}
\end{table}

\section{System Description}
In this section, we provide a comprehensive description of the system, including the system model and the associated models for UAV mobility, data transmission, and energy consumption. The system model defines the main entities involved. The models for mobility, data transmission, and energy consumption characterize the UAV’s movement, communication behavior, and energy usage, respectively, in a manner consistent with real-world operations. This modeling approach enhances the credibility of experimental results presented in Section 6. The notation used throughout the remainder of the paper is summarized in Table~\ref{table2}.

\subsection{System Model}
The system model is shown in Fig.~\ref{fig1}. The DCU collects data in area $\mathbb{A}$. The length, width, and height of $\mathbb{A}$ are $X$, $Y$, and $Z$, respectively. The system entities include the data collection UAV (DCU), other UAVs (OUs), and ground equipment (GE), which are detailed as follows.

\textbf{Data Collection UAV (DCU).} The DCU is responsible for the data collection in the low-altitude economy and flies at a fixed altitude $H$. It takes off from $\mathbb{T}\mathbb{A}$, then conducts data collection in $\mathbb{A}$, and eventually reaches $\mathbb{L}\mathbb{A}$. When conducting data collection, the DCU is not allowed to fly into NFZs. Furthermore, when the DCU flies into residential zones (RZs), to ensure noise control, its flight speed must be less than ${v_{\text{limit}}}$~\cite{28}. The DCU must also ensure that it does not collide with building zones (BZs) and other UAVs (OUs) in the low-altitude economic networking. Therefore, we assume that the DCU is equipped with a sensing device with a perception radius of $PR$. Through this device, the DCU can detect the positions and flight directions of OUs and thereby adjust its flight trajectory to avoid collisions.

\textbf{Other UAVs (OUs).} Considering the existence of airspace reuse in the low-altitude economy~\cite{36}, there will also be OUs performing other tasks in the system. These UAVs may appear on the flight trajectories where the DCU collects data. To avoid collisions, the algorithm proposed in this paper considers this issue when optimizing the DCU's flight trajectory.

\textbf{Ground Equipment (GE).} GE refers to devices that generate data in the low-altitude economy, such as intelligent traffic devices, power infrastructure equipment, and environmental monitoring devices~\cite{37}. These data are collected regularly by the DCU. $N_{\mathrm{GE}}$ GEs are placed independently and uniformly at random over $\mathbb{A}$, while maintaining appropriate clearance from obstacle boundaries and zones (NFZ/BZ/RZ). Such random deployment is widely adopted in UAV-enabled data collection trajectory planning scenarios, as it captures diverse and unpredictable real-world layouts without imposing artificial structural patterns.

\subsection{ Models Related to UAV}
This section presents models about how UAV moves, transmits data, and consumes energy in our system.

\subsubsection{Mobility Model}
The movement model is used to describe how the position of the DCU changes. The high velocity and dynamic topology of UAVs can lead to Doppler frequency shift (DFS) and fast-fading effects, which significantly degrade communication quality during data collection~\cite{41}. To mitigate these effects and maintain stable links with mobile devices, DCU adopts a \textit{hover-and-fly} strategy~\cite{42}. In this scheme, the DCU first travels within the designated area $\mathbb{A}$ to locate an optimal communication position and then hovers at that point to collect data. This hovering mechanism effectively stabilizes the communication channel by reducing DFS-induced distortion and intercarrier interference.  

To model this process in a tractable manner, the entire trajectory planning period is divided into equal time slots of length $\Delta t$, where $t \in [0, T]$ denotes the slot index and $T$ is the total number of time slots determined by mission duration. Each time slot is equally divided into a flight phase and a data collection phase. During the flight phase, the DCU moves within $\mathbb{A}$ to search for a better position, and during the data collection phase, it hovers to communicate with mobile devices. In practice, there may be cases where no devices are within the sensing range during data collection, causing idle waiting and unnecessary energy consumption.  

To minimize this energy waste and maintain modeling simplicity, the duration of both phases is set to 1~s, yielding a total slot length of $\Delta t = 2$~s. This fine-grained time division allows the DCU to frequently adjust its position while ensuring stable communication in each slot. The DCU is equipped with a single antenna and moves within the area $\mathbb{A}$. The DCU flies at a fixed altitude $H$ with a maximum speed $v_{\max}$ and can communicate with at most one device per slot.

Given the DCU’s position ${L_{\mathrm{DCU},t}} = ({x_{\mathrm{DCU},t}},{y_{\mathrm{DCU},t}},H$ and the velocity ${v_{\mathrm{DCU}}}(t) = (v_{x}(t),v_{y}(t),0)$ at the beginning of time slot $t$, the DCU’s position at time $t+1$ is as follows:
\begin{equation}
\begin{aligned}
&{L_{\mathrm{DCU},t+1}} = \\
&({x_{\mathrm{DCU},t}} + v_{x}(t){t_{\mathrm{fly}}},\ 
  {y_{\mathrm{DCU},t}} + v_{y}(t){t_{\mathrm{fly}}},\ 
  H),
\end{aligned}
\label{eq:motion1}
\end{equation}
where ${t_{\mathrm{fly}}} = \Delta t/2$ is the length of the flight phase. Given the maximum flight speed of the DCU, the distance between ${L_{\mathrm{DCU},t+1}}$ and ${L_{\mathrm{DCU},t}}$ should satisfy the following constraint:
\begin{equation}
\left\| {L_{\mathrm{DCU},t+1}} - {L_{\mathrm{DCU},t}} \right\| \le {v_{\max}}{t_{\mathrm{fly}}}.
\label{eq:motion2}
\end{equation}

If the DCU flies into ${\mathbb{R}_{\mathrm{RZ}}}$, the distance between ${L_{\mathrm{DCU},t+1}}$ and ${L_{\mathrm{DCU},t}}$ should satisfy the following constraint:
\begin{equation}
\left\| {L_{\mathrm{DCU},t+1}} - {L_{\mathrm{DCU},t}} \right\| \le {v_{{\mathrm{limit}}}}{t_{\mathrm{fly}}},
\label{eq:motion3}
\end{equation}
where $v_{\mathrm{limit}}$ is the speed limit in ${{\mathbb{R}}_{\mathrm{RZ}}}$ for the DCU.

The OUs follow the same mobility model as the DCU described in Eq.~(\ref{eq:motion1}). 
However, unlike the DCU whose trajectory is optimized by the proposed algorithm, 
the trajectories of all OUs are randomly generated prior to simulation. 
Each OU’s velocity $v_{\mathrm{OU}}(t)$ is sampled from a uniform distribution within $[0, v_{\max}]$, and its movement direction is randomly initialized and updated according to the same motion equations as the DCU. During trajectory generation, each OU’s path is constrained to ensure that it does not intersect with any BZs or NFZs, thereby maintaining physically valid and collision-free motion throughout the environment. This stochastic trajectory generation emulates the unpredictable movement of other UAVs or aerial entities in low-altitude airspace, introducing dynamic uncertainty for the DCU’s trajectory planning.

\subsubsection{Data Transmission Model}
The data transmission model describes how much the data collection volume of the DCU collects in the time slot $t$. Before giving how to compute it, we first present the effective information rate of the DCU in the time slot $t$. The DCU collects data from a single selected GE within its communication range in each time slot by using the standard time division multiple access~\cite{38}. When DCU collects data, it is in a hovering state without moving. The DCU begins to collect data at time $t + {t_{\mathrm{fly}}}$ ($t_{\mathrm{fly}} = \Delta t/2$) and lasts for ${t_{\mathrm{hover}}} = \Delta t - {t_{\mathrm{fly}}}$. As a result, assuming the GE that is currently collecting data is $\mathrm{GE}_i$, the effective information rate during the time slot $t$ is as follows:
\begin{equation}
{B_{\mathrm{GE}_i}}(t) = \log_2\left(1 + {\mathrm{SNR}}_{\mathrm{GE}_i}(t)\right),
\label{eq:rate}
\end{equation}
where ${\mathrm{SNR}}_{\mathrm{GE}_i}(t) = TP_{\mathrm{GE}_i} \cdot G_{\mathrm{GE}_i}(t) / \mathcal{N}$ is the signal-to-noise ratio at the beginning of time slot $t$, $TP_{\mathrm{GE}_i}$ is the transmit power of $\mathrm{GE}_i$, $G_{\mathrm{GE}_i}(t)$ is the channel power gain from the DCU to $\mathrm{GE}_i$ at the beginning of time slot $t$, and $\mathcal{N}$ is the noise power.

Given the fixed flight altitude of the DCU $H$, the communication link with $\mathrm{GE}_i$ is assumed to follow a line-of-sight (LoS) air-to-ground (A2G) channel model~\cite{51}. This assumption aligns well with the characteristics of UAV-assisted data collection, where the DCU actively approaches each GE and performs data acquisition at short distances during hovering. 
Such close-range outdoor A2G communication typically presents dominant LoS components, making the LoS model a widely adopted and effective abstraction in UAV data collection trajectory planning scenarios. The free-space path-loss model is used to compute $G_{\mathrm{GE}_i}(t)$ as follows~\cite{44}:
\begin{equation}
\begin{aligned}
&G_{\mathrm{GE}_i}(t) = \\
&\frac{\alpha}
{\sqrt{|| (x_{\mathrm{DCU},i + t_{\mathrm{fly}}}, y_{\mathrm{DCU},i + t_{\mathrm{fly}}}, H) - (x_{{\mathrm{GE}}_{i}}, y_{{\mathrm{GE}}_{i}}, 0) ||^2}},
\end{aligned}
\label{eq:pathloss}
\end{equation}
where $\alpha$ is the channel power gain at 
$\sqrt{\left\| (x_{\mathrm{DCU},i + t_{\mathrm{fly}}}, y_{\mathrm{DCU},i + t_{\mathrm{fly}}}, H) - (x_{{\mathrm{GE}}_{i}}, y_{{\mathrm{GE}}_{i}}, 0) \right\|^2} = 1$ m distance.

Note that, if the DCU wants to transmit data, ${\mathrm{SNR}}_{\mathrm{GE}_i}(t)$ must exceed a certain threshold $\tau$~\cite{45}. Therefore, according to Eqs.~\eqref{eq:rate} and~\eqref{eq:pathloss}, the data collection volume of the DCU collects in the time slot $t$ $TD_{\mathrm{GE}_i}(t)$ can be computed as follows:

\begin{equation}
\begin{aligned}
TD_{\mathrm{GE}_i}(t) =
\begin{cases}
B_{\mathrm{GE}_i}(t) {t_{\mathrm{hover}}}, 
& \begin{aligned}[t]
&\text{if } DV_{\mathrm{GE}_i,t} > B_{\mathrm{GE}_i}(t) \\
&\text{and } {\mathrm{SNR}}_{\mathrm{GE}_i}(t) \ge \tau,
\end{aligned} \\[6pt]

DV_{\mathrm{GE}_i,t}, 
& \begin{aligned}[t]
&\text{if } DV_{\mathrm{GE}_i,t} \le B_{\mathrm{GE}_i}(t) \\
&\text{and } {\mathrm{SNR}}_{\mathrm{GE}_i}(t) \ge \tau,
\end{aligned} \\[6pt]

0, 
& \text{if } {\mathrm{SNR}}_{\mathrm{GE}_i}(t) < \tau,
\end{cases}
\end{aligned}
\label{eq:throughput}
\end{equation}
where $DV_{\mathrm{GE}_i,t}$ is the data volume of the $i$-th GE at the beginning of time slot $t$.

\subsubsection{Energy Consumption Model}
The energy consumption model is used to describe how much the remaining energy of the the DCU changes in the time slot \(t\). The energy consumption of the DCU in the time slot \(t\) can be divided into two parts: the energy consumed by flying (denoded as \(e_{\text{fly}}(t)\)) and the energy consumed by hovering (denoted by \(e_{\text{hover}}(t)\)). For \(e_{\text{fly}}(t)\), it can be computed as follows~\cite{47}:
\begin{equation}
\begin{aligned}
&e_{\text{fly}}(t) =\\
&\Big(\frac{\lambda}{8}\rho \eta A_r \Im^3 r_{\text{rotor}}^3 \left(1 + \frac{3 v_{\mathrm{DCU}}(t)^2}{U_{\text{tip}}^2} \right) + \frac{1}{2} d_0 \rho s A_r v_{\mathrm{DCU}}(t)^3 \Big)t_{\text{fly}},
\end{aligned}
\label{eq:efly}
\end{equation}
where \(\lambda\) denotes the profile drag coefficient, \(A_r\) the rotor disc area, \(\Im\) the blade angular velocity, \(r_{\text{rotor}}\) the rotor radius, \(d_0\) denotes the fuselage drag coefficient, \(\rho\) the density of air, \(\eta\) the rotor solidity factor, \(U_{\text{tip}}\) the tip velocity of the rotor blade, and \(v_{\mathrm{DCU}}\) the velocity of DCU.

For \(e_{\text{hover}}(t)\), it can be computed as follows:
\begin{equation}
\begin{aligned}
&e_{\text{hover}}(t) = \\
&\left( (1 + \Re) w^{3/2} \sqrt{1 + \frac{v_{\mathrm{DCU}}(t)^4}{4 v_0^4}} - \frac{v_{\mathrm{DCU}}(t)^2 / 2 v_0}{\sqrt{2 \rho A_r}} \right) t_{\text{hover}},
\end{aligned}
\label{eq:ehover}
\end{equation}
where \(\Re\) is the incremental correction factor to induced power, \(w\) the weight of DCU, and \(v_0\) the mean rotor induced velocity in hover.

Note that since the energy consumed by DCU for communication is much less than that consumed for hovering~\cite{46}, when calculating \(e_{\text{hover}}(t)\), we ignore the energy consumed by DCU for data transmission. Besides, the UAV-related parameter values in this section follow the settings provided in~\cite{47}.

\section{Optimization Problem}
In this section, we present the main requirements behind the optimization problem to highlight the motivation before formally defining it. Then we detail the formalized problem and its characteristic.

\subsection{Requirement Analysis}
The optimization problem involves two primary requirements. The first requirement includes the constraints imposed by the physical environment, and the second one involves the task-specific requirements associated with data collection in the low-altitude economy. For the former, the following requirement is taken into consideration:

\textbf{R1) Limited Mobility.} The moving distance of DCU during each time slot should satisfy the Eq.~\eqref{eq:motion2}.

For the latter, the following requirements are taken into consideration:

\textbf{R2) Robust Obstacle Avoidance.} Since data collection in the low-altitude economy is concentrated in urban areas and there is airspace reuse in low-altitude areas~\cite{36}, when the DCU conducts data collection, it not only needs to avoid ${{\mathbb{R}}_{\mathrm{BZ}}}$, ${{\mathbb{R}}_{\mathrm{NFZ}}}$, but also OUs.

\textbf{R3) Compliance.} Unlike general trajectory planning tasks, DCU operating in urban low-altitude environments must comply with strict regulations. They must reduce speed near ${{\mathbb{R}}_{\mathrm{RZ}}}$ to minimize noise and meet regulatory standards.

\textbf{R4) Economy.} Within the low-altitude economy, trajectory planning tasks must jointly consider the maximization of data acquisition and the minimization of energy consumption to maximize the trajectory planning task benefits. Consequently, the DCU is required not only to collect sufficient data but also to complete the task along an energy-efficient trajectory that minimizes energy costs.

\subsection{Problem Formulation}
According to the requirement analysis introduced in Section 4.1, we try to optimize the flight trajectory of DCU while maximizing the amount of collected data. The optimization problem can be formally formulated as follows:

\begin{equation}
\textbf{OP:}\quad \underset{v_{\mathrm{DCU}}(t)}{\arg\max}\;\sum\nolimits_{t=1}^T TD_{\mathrm{GE}_i}(t)
\label{eq:op}
\end{equation}
\text{s.t.}
\begin{align} 
& \mathrm{GE}_i = \arg \max \left( \{ B_{\mathrm{GE}_i}(t) \}_{i = 1}^{N_{\mathrm{GE}}} \wedge ({\mathrm{SNR}}_{\mathrm{GE}_i}(t) \ge \tau ) \right), \tag{9a} \\
& L_{\mathrm{DCU},t+1} \nonumber \\ 
& = (x_{\mathrm{DCU},t} + v_{x}(t)t_{\text{fly}},\ y_{\mathrm{DCU},t} + v_{y}(t)t_{\text{fly}},\ H), \tag{9b} \\
& v_{\mathrm{DCU}}(t) = (v_{x}(t), v_{y}(t),0) \le v_{\max}, \quad \forall t, \tag{9c} \\
& v_{\mathrm{DCU}}(t) \le v_{\text{limit}}, \quad \text{if } L_{\mathrm{DCU},t} \in \{ \mathbb{R}_{\mathrm{RZ},\mathit{m}} \}_{m=1}^{N_{\mathrm{RZ}}}, \tag{9d} \\
&(L_{\mathrm{DCU},t}, L_{\mathrm{OU},t}) \in \mathbb{A}, \quad \forall t, \tag{9e} \\
& (L_{\mathrm{DCU},t}, L_{\mathrm{OU},t}) \notin \{ \mathbb{R}_{\mathrm{NFZ},\mathit{l}} \}_{l=1}^{N_{\mathrm{NFZ}}}, \quad \forall t, \tag{9f} \\
&(L_{\mathrm{DCU},t}, L_{\mathrm{OU},t}) \notin \{ \mathbb{R}_{\mathrm{BZ},\mathit{n}} \}_{n=1}^{N_{\mathrm{BZ}}}, \quad \forall t, \tag{9g}  \\ 
& \left\| L_{\mathrm{DCU},t} - L_{\mathrm{OU}_j,t} \right\| > 0, \quad \forall t,\ j \in [1, N_{\mathrm{OU}}], \tag{9h} \\
& (L_{\mathrm{DCU},1} = \mathrm{SL} \in \mathbb{TA}) \wedge (L_{\mathrm{DCU},T} = \mathrm{EL} \in \mathbb{LA}), \tag{9i} \\
& \sum\nolimits_{t = 1}^T (e_{\text{fly}}(t) + e_{\text{hover}}(t)) \le E_{\text{limit}}, \quad \forall t. \tag{9j}
\end{align}

The objective function \textbf{OP} maximizes the amount of data collected over the given time slots. Constraint (9a) selects a GE to collect data because the DCU follows a maximum signal strength policy to connect to the GE with the highest received signal power~\cite{51}. Note that constraint (9a) defines the feasible region for data collection, where a GE can be served only when the DCU is within its communication range, instead of enforcing a distance-priority rule. Once the DCU enters the feasible range of a GE, data collection occurs automatically. When multiple GEs have the same effective rate $B_{\mathrm{GE}_i}(t)$, a randomly GE will be collected data by the DCU.
Constraint (9b) governs the position updates of the DCU according to the mobility model. Constraint (9c) enforces the maximum speed limit, and Constraint (9d) enforces the maximum speed limit in \(\mathbb{R}_{\mathrm{RZ}}\) to satisfy \textbf{R1} and \textbf{R3}. Constraints (9e), (9f), (9g), and (9h) ensure that the DCU operates within \(\mathbb{A}\), avoids \(\mathbb{R}_{\mathrm{NFZ}}\), and prevents collisions with \(\mathbb{R}_{\mathrm{BZ}}\) or OUs to satisfy \textbf{R2} and \textbf{R3}. constraint (9i) ensures DCU takes off from $\mathbb{TA}$ and lands off in $\mathbb{LA}$, while constraint (9j) ensures the maximum energy consumption of the DCU in a trajectory planning task to satisfy \textbf{R4}.

\subsection{Problem Characteristics}

This optimization problem involves continuous control, nonlinear constraints, dynamic environmental factors (such as OUs), as well as regulatory and energy limitations, making it structurally complex and difficult. Traditional optimization algorithms often perform inefficiently and lack adaptability when dealing with such high-dimensional, dynamic, and partially observable problems. In contrast, DRL offers adaptive learning capabilities, making it well-suited for continuous control and long-term optimization. LLMs, on the other hand, can provide rule-compliant and reasoning decisions in critical scenarios by leveraging rich prior knowledge. The integration of DRL and LLM leverages efficiency, generalization, and safety, making it more suitable than traditional methods to solve complex UAV trajectory planning problems.

\section{Proposed Algorithm}
This section first formulates the optimization problem from Section 4 as a POMDP to enable solution via DRL. We then present SAC algorithm, which serves as the foundation for our algorithm. Finally, we introduce the proposed algorithm and demonstrate its effectiveness in addressing the formulated problem.

\subsection{POMDP Modeling}
As the future position of the DCU is determined solely by its current state (i.e., position and velocity), the trajectory planning problem satisfies the Markov property. Nevertheless, due to the DCU’s limited ability to perceive the full environmental state, the problem is more appropriately formulated as a POMDP, defined by a corresponding tuple \((\mathcal{O},\mathcal{A},r,\mathcal{P},\delta )\), where the observation space \(\mathcal{O}\) replaces the state space~\cite{48}. Then, the details of \(\mathcal{O}\), \(\mathcal{A}\), and \(r\) in \((\mathcal{O},\mathcal{A},r,\mathcal{P},\delta )\) are as follows:

\textbf{Observation Space \(\mathcal{O}\)}: \(o_t \in \mathcal{O}\) is the state that the DCU can observe at time \(t\), where
$
o_t = (o_{\mathrm{GE},t}, o_{Z,t}, o_{\mathrm{OU},t}, o_{\mathrm{DCU},t}).
$
\begin{itemize}
  \item \(o_{\mathrm{GE},t}\) is the state about all GEs and \(o_{\mathrm{GE},t} = \{ L_{{\mathrm{GE}}_i}, DV_{\mathrm{GE}_i,t}, TP_{{\mathrm{GE}}_{i}} \}_{i=1}^{N_{\mathrm{GE}}}\).
  \item \(o_{Z,t}\) is the state about all zones and \(o_{Z,t} = \{ \{ \mathbb{R}_{\mathrm{NFZ},l} \}_{l=1}^{N_{\mathrm{NFZ}}}, \{ \mathbb{R}_{\mathrm{RZ},m} \}_{m=1}^{N_{\mathrm{RZ}}}, \{ \mathbb{R}_{\mathrm{BZ},n} \}_{n=1}^{N_{\mathrm{BZ}}} \}\).
  \item  $o_{\mathrm{OU},t}$ represents the state information of OUs located within the perception radius of the DCU during time slot $t$. It is defined as $o_{\mathrm{OU},t} = \{ (L_{\mathrm{OU}_j,t}, v_{\mathrm{OU}_j,t}) \}_{j=1}^{N_{\mathrm{OU}}}$, where $L_{\mathrm{OU}_j,t}$ and $v_{\mathrm{OU}_j,t}$ denote the position and velocity of $\mathrm{OU}_j$ at the beginning of time slot $t$. The number of OUs within the perception radius of the DCU may vary across different time slots, resulting in changes in the dimension of $o_{\mathrm{OU},t}$. To maintain a consistent input size, the zero-padding method is employed to standardize the dimension of $o_{\mathrm{OU},t}$.
  \item \(o_{\mathrm{DCU},t}\) is the state about DCU itself and \(o_{\mathrm{DCU},t} = \{ L_{\mathrm{DCU},t}, v_{\mathrm{DCU},t}, E(t), TD_{\mathrm{GE}_i}(t), \mathbb{T}\mathbb{A}, \mathbb{L}\mathbb{A} \}\).
\end{itemize}

\textbf{Action Space \(\mathcal{A}\)}: \(a_t \in \mathcal{A}\) is the sampled action at time \(t\) and $a_t = (v_x(t), v_y(t)),$ which means the DCU will move at the velocity \((v_x(t), v_y(t))\) at time \(t\).

\textbf{Reward function \(r\)}: \(r\) is used to effectively guide the trajectory planning during the data collection process and plays a critical role in both the efficiency and performance of policy learning. To facilitate learning an optimal policy aligned with the objective function \textbf{OP} and its associated constraints, we formulate a composite reward function, expressed as follows:
\begin{equation}
   r_t = r_{1t} + r_{2t} + r_{3t} + r_{4t} + r_{5t} + r_{6t} + r_{7t} + r_{8t} + r_{9t}.
\label{eq:r} 
\end{equation}


${r_{1t}}$ encourages the DCU to collect data, and is as follows:
\begin{equation}
    {r_{1t}} = \left\{
    \begin{array}{ll}
    {\sigma_1}, & \text{if } TD_{\mathrm{GE}_i}(t + 1) \ge 0, \\
    0, & \text{otherwise}.
    \end{array}
\right.
\end{equation}

${r_{2t}}$ is used to ensure that the DCU will not collide with OUs and is as follows:
\begin{equation}
r_{2t} = \left\{
\begin{array}{ll}
0, & \text{if } d_{t,\mathrm{OU}} \ge d_{\mathrm{safe},\mathrm{OU}}, \\
- \sigma_2 \displaystyle 
\frac{d_{\mathrm{safe},\mathrm{OU}} - d_{t,\mathrm{OU}}}
     {d_{\mathrm{safe},\mathrm{OU}} - d_{\min,\mathrm{OU}}}, 
& \begin{aligned}[t]
& \text{if } d_{\min,\mathrm{OU}} < d_{t,\mathrm{OU}}. \\[-0.8ex]
& \text{and } d_{t,\mathrm{OU}} < d_{\mathrm{safe},\mathrm{OU}},
\end{aligned} \\[0.8ex]
- \sigma_2, & \text{if } d_{t,\mathrm{OU}} \le d_{\min,\mathrm{OU}},
\end{array}
\right.
\end{equation}
where ${d_{t,\mathrm{OU}}} = \arg\min \{ ||L_{\mathrm{DCU},t} - L_{\mathrm{OU}_j,t}|| \}_{j = 1}^{N_{\mathrm{OU}}}$, ${d_{\min ,\mathrm{OU}}}$ is the minimum safe distance between the DCU and OUs, which is used for the DCU to adjust the speed to avoid collisions, and ${d_{\mathrm{safe},\mathrm{OU}}}$ is the safe distance threshold between the DCU and OUs.

${r_{3t}}$ is used to ensure that the DCU will not collide with tall buildings and is as follows:
\begin{equation}
\begin{aligned}
r_{3t} = 
\begin{cases}
0, 
& \text{if } d_{t,\mathrm{BZ}} \ge d_{\mathrm{safe},\mathrm{BZ}}, \\

- \sigma_3 \times 
\frac{d_{\mathrm{safe},\mathrm{BZ}} - d_{t,\mathrm{BZ}}}{d_{\mathrm{safe},\mathrm{BZ}} - d_{\min ,\mathrm{BZ}}}, 
& \begin{aligned}[t]
& \text{if } d_{\min ,\mathrm{BZ}} < d_{t,\mathrm{BZ}} \\
& \text{and } d_{t,\mathrm{BZ}} < d_{\mathrm{safe},\mathrm{BZ}},
\end{aligned} \\

- \sigma_3, 
& \text{if } d_{t,\mathrm{BZ}} \le d_{\min ,\mathrm{BZ}},
\end{cases}
\end{aligned}
\end{equation}
where ${d_{t,\mathrm{BZ}}} = \arg\min \{ ||L_{\mathrm{DCU},t} - \mathbb{R}_{\mathrm{BZ},n}|| \}_{n = 1}^{N_{\mathrm{BZ}}}$, ${d_{\min ,\mathrm{BZ}}}$ is the minimum safe distance between the DCU and tall buildings, which is used for the DCU to adjust the speed to avoid collisions, and ${d_{\mathrm{safe},\mathrm{BZ}}}$ is the safe distance threshold between the DCU and tall buildings.

${r_{4t}}$ is used to ensure that the DCU will not fly into NFZs and is as follows:
\begin{equation}
\begin{aligned}
& r_{4t} = \\
&\begin{cases}
0, 
& \text{if } d_{t,\mathrm{NFZ}} \ge d_{\mathrm{safe},\mathrm{NFZ}}, \\

- \sigma_4 \times 
\frac{d_{\mathrm{safe},\mathrm{NFZ}} - d_{t,\mathrm{NFZ}}}{d_{\mathrm{safe},\mathrm{NFZ}} - d_{\min ,\mathrm{NFZ}}}, 
& \begin{aligned}[t]
& \text{if } d_{\min ,\mathrm{NFZ}} < d_{t,\mathrm{NFZ}} \\
& \text{and } d_{t,\mathrm{NFZ}} < d_{\mathrm{safe},\mathrm{NFZ}},
\end{aligned} \\

- \sigma_4, 
& \text{if } d_{t,\mathrm{NFZ}} \le d_{\min ,\mathrm{NFZ}},
\end{cases}
\end{aligned}
\end{equation}

\noindent
where $d_{t,\mathrm{NFZ}} = \arg\min \left\{ \| L_{\mathrm{DCU},t} - \mathbb{R}_{\mathrm{NFZ},l} \| \right\}_{l = 1}^{N_{\mathrm{NFZ}}}$, $d_{\min ,\mathrm{NFZ}}$ is the minimum safe distance between the DCU and NFZs, which is used for the DCU to adjust the speed to avoid collisions, and $d_{\mathrm{safe},\mathrm{NFZ}}$ is the safe distance threshold between the DCU and NFZs.

${r_{5t}}$ is used to ensure that the speed of the DCU when flying into residential areas meets the speed limit and is as follows:
\begin{equation}
\begin{aligned}
r_{5t} = 
\begin{cases}
0, 
& \text{if } v_{\mathrm{DCU}}(t) \le v_{\text{limit}} 
\text{ and } \\
&L_{\mathrm{DCU},t} \in \left\{ \mathbb{R}_{\mathrm{RZ},m} \right\}_{m = 1}^{N_{\mathrm{RZ}}}, \\

- \sigma_5 \times 
\frac{v_{\mathrm{DCU}}(t) - v_{\text{limit}}}{v_{\max} - v_{\text{limit}}}, 
& \begin{aligned}[t]
& \text{if } v_{\text{limit}} < v_{\mathrm{DCU}}(t) \le v_{\max} \\
& \text{and } L_{\mathrm{DCU},t} \in \left\{ \mathbb{R}_{\mathrm{RZ},m} \right\}_{m = 1}^{N_{\mathrm{RZ}}}.
\end{aligned}
\end{cases}
\end{aligned}
\end{equation}

${r_{6t}}$ is used to ensure that the DCU should not fly out of the given area and is as follows:
\begin{equation}
\begin{aligned}
r_{6t} =
\begin{cases}
0, 
& \text{if } d_{t,\mathbb{A}} \ge d_{\mathrm{safe},\mathbb{A}}, \\[6pt]
- \sigma_6 \times \dfrac{d_{\mathrm{safe},\mathbb{A}} - d_{t,\mathbb{A}}}{d_{\mathrm{safe},\mathbb{A}} - d_{\min ,\mathbb{A}}}, 
& \text{if } d_{\min ,\mathbb{A}} < d_{t,\mathbb{A}} < d_{\mathrm{safe},\mathbb{A}}, \\[6pt]
- \sigma_6, 
& \text{if } L_{\mathrm{DCU},t} \notin \mathbb{A},
\end{cases}
\end{aligned}
\end{equation}

\noindent
where $d_{t,\mathbb{A}} = \| L_{\mathrm{DCU},t} - \mathbb{A} \|$ is the distance between the DCU and the boundary of $\mathbb{A}$ in the time slot $t$, $d_{\min ,\mathbb{A}}$ is the minimum safe distance between the DCU and the boundary of $\mathbb{A}$, which is used for the DCU to adjust the speed to avoid flying out, and $d_{\mathrm{safe},\mathbb{A}}$ is the safe distance threshold between the DCU and the boundary of $\mathbb{A}$.

$r_{7t}$ is used to encourage that the DCU can land within the landing area and is as follows:
\begin{equation}
\begin{aligned}
r_{7t} =
& \begin{cases}
0, 
& \text{if } d_{t,\mathbb{LA}} \ge d_{tar,\mathbb{LA}}, \\[6pt]
- \sigma_7 \times \left( 1 - \dfrac{d_{t,\mathbb{LA}}}{d_{tar,\mathbb{LA}}} \right), 
& \text{if } 0 < d_{t,\mathbb{LA}} < d_{tar,\mathbb{LA}}.
\end{cases}
\end{aligned}
\end{equation}

$r_{8t}$ is used to encourage the DCU to return to the landing area before its energy is exhausted and is as follows:
\begin{equation}
\begin{aligned}
r_{8t} = 
& \begin{cases}
0, 
& \text{if } E(t) \ge E_{\min}(t), \\[6pt]
- \sigma_8 \times \dfrac{E_{\min}(t) - E(t)}{E_{\min}(t)}, 
& \text{if } E(t) < E_{\min}(t),
\end{cases}
\end{aligned}
\end{equation}

\noindent
where $E_{\min}(t)$ is the minimum energy required for DCU to reach the $\mathbb{LA}$ at the beginning of time slot $t$ and $E_{\min}(t) = e_{\text{fly}}(t) \cdot d_{t,\mathbb{LA}} / v_{\max}$.

$r_{9t}$ is used to encourage the DCU can complete the data collection quickly to reduce energy consumption and is as follows:
\begin{equation}
r_{9t} = - \sigma_9.
\end{equation}

\subsection{SAC Algorithm}
SAC is an off-policy DRL method based on the maximum-entropy framework~\cite{39,40}. 
Its off-policy property enables efficient training and easy incorporation of LLM guidance, 
making it suitable for continuous-control problems such as UAV trajectory planning.

SAC maximizes both the expected cumulative reward and the policy entropy:
\begin{equation}
\pi^* = \arg \max_{\pi} 
\sum_{t=0}^{\infty} \mathbb{E}_{\mathcal{P},\pi}
\!\left[\delta^t \big(r_t + \mu\,\mathcal{H}(\pi(\cdot\mid o_t))\big)\right],
\end{equation}
where $\delta$ is the discount factor, $\mu$ is the temperature parameter, and 
$\mathcal{H}(\pi(\cdot\mid o_t)) = -\int \pi(a_t\mid o_t)\log\pi(a_t\mid o_t)\,da_t$ 
is the policy entropy. The soft Bellman equation is defined as:
\begin{equation}
\left\{
\begin{aligned}
Q_{\text{soft}}(o_t,a_t) &= r_t + \delta\,\mathbb{E}_{\mathcal{P}}\!\left[V_{\text{soft}}(o_{t+1})\right],\\
V_{\text{soft}}(o_t) &= \mathbb{E}_{a_t\sim\pi}\!\left[Q_{\text{soft}}(o_t,a_t) - \mu \log \pi(a_t\mid o_t)\right].
\end{aligned}
\right.
\end{equation}

The critic network minimizes the following loss:
\begin{equation}
J_Q(\varphi_i)=
\mathbb{E}_{(o_t,a_t,r_t,o_{t+1})\sim D}
\!\left[\big(Q_i^{\theta}(o_t,a_t)-y_t\big)^2\right],
\end{equation}
with target value 
$y_t=r_t+\delta\!\left(\min_{j}Q_j^{\theta_{\text{tar}}}(o_{t+1},\tilde a_{t+1})
-\mu\log\pi^{\phi}(\tilde a_{t+1}\mid o_{t+1})\right)$,
where $D$ is the experience pool, and $\theta_{\text{tar}}$ denotes target critic parameters. The actor is optimized via reparameterization:
\begin{equation}
\begin{aligned}
J_\pi(\phi)=
\mathbb{E}_{o_t\sim D,\,\varsigma\sim\mathcal{K}(0,1)}
\Big[&\,\mu\log\pi^\phi(\tilde{a}_t^\phi(o_t,\varsigma)\mid o_t)\\
&-Q^{\theta}(o_t,\tilde{a}_t^\phi(o_t,\varsigma))\Big],
\end{aligned}
\end{equation}
where $\tilde{a}_t^\phi(o_t,\varsigma)
=\tanh\!\big(\beta^\phi(o_t)+\alpha^\phi(o_t)\odot\varsigma\big)$,
$\beta^\phi$ and $\alpha^\phi$ denote the Gaussian mean and standard deviation, 
$\varsigma$ is a standard normal noise, and $\odot$ is element-wise multiplication.

Through iterative updates of $\pi$, $Q$, and $V$, SAC achieves a stable, 
high-performance policy for continuous control under uncertainty.

\begin{figure*}[!t]
    \centering
    \captionsetup{justification=raggedright, singlelinecheck=false}
    \includegraphics[width=5.0in]{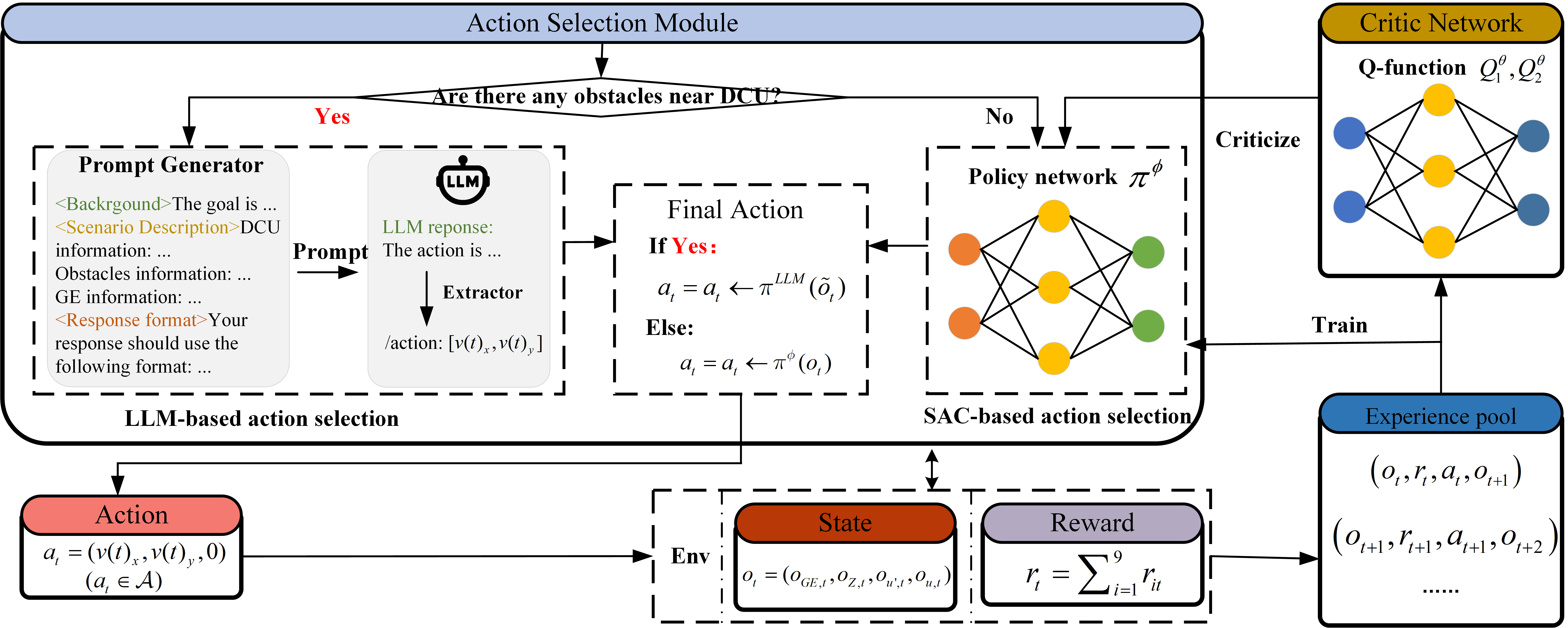}
    \caption{The proposed algorithm integrates an LLM into SAC for conditional action selection. The LLM is invoked to generate actions when obstacles are detected; otherwise, actions are selected by the SAC policy network. The chosen actions interact with the environment, and transitions are stored to train the critic networks.}
    \label{fig2}
    \vspace{-10pt}
\end{figure*}

\subsection{Our Algorithm}
This section presents the core design principles of our trajectory planning algorithm, outlines its detailed implementation, and provides a comprehensive analysis of its computational complexity.

\subsubsection{Main Idea}
Based on the POMDP formulation in Section 5.1 and the continuous nature of DCU’s velocity, we adopt SAC algorithm to optimize the DCU’s flight trajectory during the data collection. However, in the context of low-altitude economy scenarios, where strict regulatory compliance and robust obstacle-avoidance capability are critical, the use of DRL alone may be insufficient due to its reliance on trial-and-error learning and lack of prior domain knowledge.
To address this issue, we propose an enhanced decision-making algorithm by integrating an LLM into the action selection module, as illustrated in Fig. ~\ref{fig2}. Specifically, when the DCU detects the presence of nearby obstacles (e.g., BZs, NFZs, RZs, or OUs), a decision condition is triggered that activates the LLM-based action selection pathway. In this mode, a \textbf{Prompt Generator} compiles semantic information, including environmental context, DCU status, and obstacle details, into a structured prompt, which is then input to an LLM. The LLM responds with a recommended action, leveraging its rich prior knowledge to ensure safety and regulatory compliance. An \textbf{Extractor} then parses the LLM’s response into actionable control parameters for the DCU.
Conversely, when no immediate obstacle is detected, the system defaults to the standard SAC policy network for action selection. This hybrid structure enables the agent to benefit from both data-driven learning and knowledge-driven reasoning: the SAC component provides efficient trajectory planning in general scenarios, while the LLM supplements decision-making in critical or unfamiliar environments requiring sophisticated judgment.
By combining the generalization capability of DRL with the reasoning capability and prior knowledge of an LLM, the proposed algorithm achieves superior performance in complex real-world scenarios, ensuring not only energy efficiency but also operational safety and compliance.

\begin{figure}[t]
\centering
\begin{tikzpicture}[every node/.style={inner sep=0pt,outer sep=0pt,scale=0.95}]
  \def\W{0.95\columnwidth}
  \def\Ysep{0cm} 

  \node (A) [anchor=north west]
  {\begin{tcolorbox}[enhanced,
      sharp corners,
      boxrule=0.4pt,
      colback=boxblue,
      colframe=boxblueframe,
      left=2pt,right=2pt,top=2pt,bottom=2pt,
      width=\W,fontupper=\scriptsize]
  (a) Task definition and observation description: you are a reinforcement learning decision agent controlling a Data Collection UAV (DCU) in a 500×500 m low-altitude environment. 
  The primary objective is to ensure safety and avoid collisions. 
  The secondary objective is to maintain compliance—never enter NFZs, and if inside an RZ, do not exceed the local speed limit 5m/s.  
  It can observe all GEs and their remaining data, static obstacles (buildings, NFZs, RZs), and only the nearby dynamic obstacles (OUs) within the observation radius. 
  Decide the next continuous velocity vector $(v_x, v_y)$ for the DCU. 
  The observation includes the DCU state (position, velocity), GEs (positions and remaining data), static obstacles (positions of buildings, NFZs, and RZs), and dynamic obstacles (visible OUs and positions).
  \end{tcolorbox}};

  \node (B) [anchor=north west, yshift=-\Ysep] at (A.south west)
  {\begin{tcolorbox}[enhanced,
      sharp corners,
      boxrule=0.4pt,
      colback=boxred,
      colframe=boxredframe,
      left=2pt,right=2pt,top=2pt,bottom=2pt,
      width=\W,fontupper=\scriptsize]
  (b) Chain-of-thought reasoning guidelines: 
  1) Safety first: ensure $(v_x, v_y)$ will not cause collisions with buildings, NFZ borders, or visible OUs; 
  2) Compliance: never enter NFZs and within RZs ensure $|v| \leq \text{rz\_limit}$; 
  3) Data efficiency: move toward GEs, which have remaining data; 
  4) Energy efficiency: prefer smaller $|v|$ if equally effective; 
  5) Stability: maintain current direction when options are equivalent.
  \end{tcolorbox}};

  \node (C) [anchor=north west, yshift=-\Ysep] at (B.south west)
  {\begin{tcolorbox}[enhanced,
      sharp corners,
      boxrule=0.4pt,
      colback=boxyellow,
      colframe=boxyellowframe,
      left=2pt,right=2pt,top=2pt,bottom=2pt,
      width=\W,fontupper=\scriptsize]
  (c) Action space and output format: the action is a continuous 2D velocity $(v_x, v_y)$, each within $[-10, 10]$. 
  The magnitude $|v| = \sqrt{v_x^2 + v_y^2}$ represents the flight speed (m/s), where +x denotes east and +y denotes north. 
  The model must output a strict JSON object as follows: \\
  \texttt{\scriptsize\{"vx":4.2,"vy":1.6,"Confidence":0.86,"Reasoning":"Moves toward GE with remaining data, avoids NFZ/OUs/buildings, and respects RZ."\}}
  \end{tcolorbox}};
\end{tikzpicture}

\caption{CoT-based prompt generator used for UAV control. 
Part (a) defines the task and observation, (b) describes the reasoning principles, and (c) specifies the action format and compact JSON output.}
\label{fig:cot_prompt_generator}
\end{figure}

\subsubsection{Algorithm Detail}
Our proposed algorithm, which is detailed in \textbf{Algorithm 1}, is designed to enable the DCU to effectively perform trajectory planning by learning an optimal velocity control policy that guides its trajectory to maximize data acquisition and avoid collision. The main process of our algorithm is shown in Fig.~\ref{fig2}.

The proposed algorithm takes the environment as input (\textbf{Line~1}) and outputs an optimal policy network (\textbf{Line~2}) that governs the DCU velocity control. Upon initialization (\textbf{Line~3}), the initialization of the Q-function parameters, policy network parameters, the LLM, the experience pool, and the target network parameters is conducted.

For each training episode, the environment is first reset (\textbf{Line~5}). Then, at each time step within the episode, the algorithm observes the current state and determines whether to invoke the LLM for action selection (\textbf{Lines~7--8}). It is important to note that this state (\(\tilde{o}_t\)) differs from the state (\(o_t\)) used by the policy network. Specifically, it refers to the pre-processed, semantically rich environmental information, such as the positions of obstacles within the DCU’s observation range, as illustrated in Fig.~\ref{fig2}. If the decision condition is met, i.e., there are obstacles near the DCU, the LLM is utilized to select an action for the DCU based on the semantic state information generated by the prompt generator. Otherwise, the action is selected using the policy network (\textbf{Lines~9--12}). The selected action is then executed, resulting in the next state and corresponding reward for the subsequent time slot. Simultaneously, the current state, action, and reward are stored in the experience pool (\textbf{Lines~14--15}). The algorithm then checks whether the current episode has terminated. If so, it proceeds to the next episode (\textbf{Line~17}).

For each gradient update step, a batch of samples is drawn from the experience pool (\textbf{Line~21}). Based on this batch, the target values are computed, and the Q-function, policy network, and target network parameters are updated accordingly (\textbf{Lines~22--24}). This process repeats until the episode concludes, after which the next episode begins. Upon completion of all training episodes, the algorithm outputs the final optimized policy network.

After the hybrid SAC–LLM training process described above, we further detail the design of the LLM \textbf{Prompt Generator}, which enables the integration of LLM-based reasoning into the decision loop. 
When the decision condition is triggered, the prompt generator converts the UAV’s structured observation into a semantically rich text prompt that guides the LLM’s inference. 
As shown in Fig.~\ref{fig:cot_prompt_generator}, the generator follows a chain-of-thought (CoT) design composed of three tightly connected parts: 
(i) task definition and observation description, which specifies the UAV’s operating environment and observation elements; 
(ii) reasoning guidelines, which instruct the LLM to reason according to safety, compliance, and efficiency rules; and 
(iii) output format, which constrains the model to produce a strict JSON object representing a continuous velocity vector $(v_x, v_y)$, confidence score, and reasoning summary. 
This prompt structure ensures interpretability and physical validity of the LLM’s output, enabling seamless conversion into executable control actions for the UAV. Additionally, to address the potential issue of hallucination by LLM, which can lead to the failure of producing valid actions, an \textbf{Extractor} is employed. The extractor first checks whether any action is produced. If no action is generated, the LLM is called again, with a threshold of 10 attempts. If this threshold is exceeded, the episode is terminated. Furthermore, after receiving the action, the extractor verifies whether the magnitude of the velocity vector is less than $v_{\max}$. If the velocity exceeds $v_{\max}$, the LLM is called again, with the same 10-attempt threshold. If the limit is exceeded again, the episode is terminated. This process ensures the robustness of the decision-making loop and maintains valid and feasible UAV control actions.

\begin{algorithm}[htbp]
\caption{SAC with LLM-based Trajectory Planning Algorithm}
\begin{algorithmic}[1]
\State \textbf{Input:} The environment $\mathcal{E}$
\State \textbf{Output:} The optimal policy network $\pi^*$
\State \textbf{Initialize:} Q-function parameters $\theta_1, \theta_2$, policy parameter $\phi$, the LLM $\pi^{LLM}$, experience pool $D$, target parameters $\theta_{1,\text{tar}}, \theta_{2,\text{tar}}$
\For{$\text{episode} = 1, 2, \ldots, L$}
    \State Initialize environment: $\mathcal{E} \leftarrow \text{EnvReset()}$
    \For{each $\text{step} = 1, 2, \ldots, N$}
        \State Observe state: $o_t \leftarrow \text{ObsState}(\mathcal{E})$
        \State Judge if using the LLM: $\text{bool} \leftarrow \text{useLLM}(\tilde{o}_t)$
        \If{bool}
            \State Select action from the LLM: $a_t \leftarrow \pi^{LLM}(\tilde{o}_t)$
        \Else
            \State Select action from policy network: $a_t \leftarrow \pi^{\phi}(o_t)$
        \EndIf
        \State Execute action and get new state and reward: $(o_{t+1}, r_t, \text{done}) \leftarrow \text{ExeAct}(a_t)$
        \State Store transition in experience pool: $(o_t, r_t, a_t, o_{t+1}) \to D$
        \If{done}
            \State Reset the environment and continue
        \EndIf
    \EndFor
    \For{each gradient step}
        \State Randomly sample a batch: $(o_t, r_t, a_t, o_{t+1}) \leftarrow D$
        \State Compute targets and update Q-functions: $Q_1^{\theta}, Q_2^{\theta}$
        \State Update policy network $\pi^{\phi}$
        \State Update target parameters: $\theta_{1,\text{tar}}, \theta_{2,\text{tar}}$
    \EndFor
\EndFor
\end{algorithmic}
\end{algorithm}

\subsubsection{Algorithm Computational Complexity Analysis}
During training, our algorithm largely follows the SAC framework for policy optimization. However, unlike standard SAC, our algorithm intermittently replaces the policy network's action outputs with those generated by an external LLM based on specific conditions. As a result, the overall training complexity consists of two components: the standard SAC update steps and the LLM inference overhead incurred during action selection. Due to each SAC training step including policy network updates with complexity $O(d_{\pi})$, Q-network updates with complexity $O(d_{Q})$, and target network updates with negligible computational cost, the training computation complexity of SAC is 
$O(n \cdot (d_{\pi} + 2d_{Q})),$
where $n$ is the batch size~\cite{52}. Assuming the LLM is queried for $p \cdot n$ samples per batch, the training computation complexity of the LLM is 
$O(p \cdot n \cdot L_{\text{LLM}} \cdot d_{\text{LLM}}^2),$
where $L_{\text{LLM}}$ is the number of tokens and $d_{\text{LLM}}$ is the LLM hidden layer size~\cite{54}. Therefore, the training computation complexity of our algorithm is 
$O((1-p) \cdot n \cdot (d_{\pi} + 2d_{Q}) + p \cdot n \cdot L_{\text{LLM}} \cdot d_{\text{LLM}}^2).$

It is important to clarify that the LLM is employed only during the offline training phase to assist the DRL process, rather than during real-time deployment. 
Specifically, the LLM provides high-level semantic guidance to the SAC agent, helping it select more informative and regulation-compliant actions during exploration, thereby accelerating policy convergence and improving training stability. 
Once the SAC policy network is fully trained, the LLM is completely removed from the control loop. 
During online execution, the UAV relies solely on the lightweight SAC policy for real-time trajectory planning and obstacle avoidance. 
This design ensures that the runtime control algorithm operates efficiently and achieves collision-free navigation without invoking any LLM inference, thereby fully guaranteeing real-time performance and computational feasibility.

\section{Experiments}
In this section, we present the experimental results of the proposed algorithm. Prior to that, we describe the experimental setup and evaluation metrics.

\subsection{Experiment Setting}
For the hardware, all experiments in this study were conducted on a computing platform equipped with Intel® Xeon® Gold 6230 CPUs and NVIDIA GeForce RTX 4090 GPUs.

\begin{table}[htbp]
\caption{Environment Parameters Setting}
\centering
\scriptsize
\setlength{\tabcolsep}{4pt}
\begin{tabular}{ccc|ccc|ccc}
\hline
\textbf{Para.} & \textbf{Value} & & \textbf{Para.} & \textbf{Value} & & \textbf{Para.} & \textbf{Value} & \\
\hline
$N_{\mathrm{GE}}$ & [11, 15] & & $v_{\max}$ & 10 m/s & & $\mathcal{N}$ & $10^{-6}$ & \\
$DV_{\mathrm{GE}_i}$ & [1, 3] & & $v_{\text{limit}}$ & 5 m/s & & \textit{PR} & 20 m & \\
$TP_{{\mathrm{GE}}_{i}}$ & 0.01 W & & $E_{\text{total}}$ & $1 \times 10^6$ J & & $d_{\min}, d_{\text{safe}}$ & 5 m, 15 m & \\
$N_{\mathrm{OU}}$ & [3, 7] & & $E_{\text{limit}}$ & $8 \times 10^5$ J & & $\tau$ & 3.6 & \\
\hline
\end{tabular}
\label{table3}
\end{table}

\subsubsection{Environment Parameters}
The experimental environment is simplified to a two-dimensional space with $\mathbb{A}$ of 500 m in both length and width. $\mathbb{T}\mathbb{A}$ and $\mathbb{L}\mathbb{A}$ are square regions measuring $25 \times 25$ m, located at the top-left and bottom-right corners of $\mathbb{A}$, respectively. Given the urban deployment scenario, the environment includes rectangular ${{\mathbb{R}}_{\mathrm{NFZ}}}$, ${{\mathbb{R}}_{\mathrm{RZ}}}$, and ${{\mathbb{R}}_{\mathrm{BZ}}}$ with the range of [50, 200] m for both length and width. The number of such regions are randomly chosen from the range [1, 3]. The weights assigned to the components of the reward function are set to $[8,\ 0.4,\ 0.4,\ 0.4,\ 0.6,\ 0.4,\ 0.6,\ 1,\ 0.01]$ and can be flexibly adjusted to reflect different task priorities. Other environmental parameters are detailed in Table~\ref{table3}.

\subsubsection{Algorithm Parameters}
For the algorithm, hyperparameters used by the proposed algorithm and other baselines, including SAC+LLM, SAC, Proximal Policy Optimization (PPO)~\cite{55}, DDPG~\cite{56}, and Constrained Policy Optimization (CPO)~\cite{achiam2017cpo}, are shown in Table~\ref{table4}. For the LLM used in our algorithm and other baselines, including LLM and SAC+LLM, we choose the DeepSeek-R1, which has a total of 672B parameters and 37B activated parameters~\cite{49}.

\begin{table}[htbp]
\centering
\caption{Hyperparameters Setting}
\begin{tabular}{l c}
\toprule
\textbf{Parameter} & \textbf{Value} \\
\midrule
Learning rate of actor network & 0.0001 \\
Learning rate of critic network & 0.0003 \\
Number of episodes & 4000 \\
Clip range & 0.2 \\
Batch size & 256 \\
Discount factor & 0.99 \\
Number of hidden layers of DNNs & 2 \\
Hidden layer size & 64 \\
\bottomrule
\end{tabular}
\label{table4}
\end{table}

\subsection{Evaluation Metrics}
For the metrics used to evaluate the proposed algorithm and other baselines, we use the metrics commonly adopted in general UAV data collection trajectory planning tasks~\cite{50}. In addition, we include metrics that reflect the specific requirements in this paper. All metrics are directly aligned with the core demands of urban low-altitude data collection in the emerging low-altitude economy, where robust obstacle avoidance (\textbf{R2}), compliance (\textbf{R3}), and economy (\textbf{R4}), mentioned in Section 4.1, are essential. Assuming $N_{\text{total}}$ is the number of experimental tests:

\begin{itemize}
  \item \textbf{Data collection rate (DCR):} This metric quantifies the average ratio of successfully collected data by the DCU relative to the total target data volume in $N_{\text{total}}$ data collection:
  \begin{equation}
    \text{DCR} = \frac{\sum\nolimits_{i=1}^{N_{\text{test}}} D_{\text{DCU},i}}{\sum\nolimits_{i=1}^{N_{\text{test}}} D_{\text{total},i}},
  \end{equation}
  where $D_{\text{DCU},i}$ is the collected data volume and $D_{\text{total},i}$ is the total target data volume in the $i$-th task.

  \item \textbf{Collision rate (CR):} This metric quantifies the ratio of trajectory planning tasks in which the DCU collides with OUs, $\mathbb{R}_{\mathrm{BZ}}$, or flies into $\mathbb{R}_{\mathrm{NFZ}}$:
  \begin{equation}
    \text{CR} = \frac{N_C}{N_{\text{total}}},
  \end{equation}
  where $N_C$ is the number of tasks involving collisions or regulatory violations.

  \item \textbf{Successful landing rate (SLR):} This metric quantifies the ratio of trajectory planning tasks in which the DCU can return to $\mathbb{LA}$ before depleting its energy:
  \begin{equation}
    \text{SLR} = \frac{N_{\text{SLR}}}{N_{\text{total}}},
  \end{equation}
  where $N_{\text{SLR}}$ is the number of tasks where the DCU successfully lands.

\end{itemize}

In addition to the commonly used metrics mentioned above, the following indicators are also employed to evaluate the effectiveness of satisfying the requirement for data collection trajectory planning under the low-altitude economy.

\begin{itemize}

  \item \textbf{Regulatory violation rate (RVR):} This metric quantifies the ratio of tasks in which the DCU exceeds speed limits within $\mathbb{R}_{\mathrm{RZ}}$:
  \begin{equation}
    \text{RVR} = \frac{N_{\text{RVR}}}{N_{\text{total}}},
  \end{equation}
  where $N_{\text{RVR}}$ is the number of such violations.

  \item \textbf{Energy consumption rate (ECR):} This metric quantifies the average ratio of energy consumed by the DCU relative to the total energy:
  \begin{equation}
    \text{ECR} = \frac{\sum\nolimits_{i=1}^{N_{\text{test}}} E_{\text{ECR},i}}{\sum\nolimits_{i=1}^{N_{\text{test}}} E_{\text{total},i}},
  \end{equation}
  where $E_{\text{ECR},i}$ is the consumed energy and $E_{\text{total},i}$ is the total available energy in the $i$-th task.
\end{itemize}

\subsection{Experiment Results}
In this section, we present a comparative analysis between the proposed algorithm and several baseline methods in terms of the DCU flight trajectories and the evaluation metrics mentioned above. The parameter settings of these baselines are introduced in Section 6.1.2. The baselines are as follows:

\begin{itemize}
  \item \textbf{Heuristic method:} A greed-based policy\footnote{We referenced the code from: \url{https://github.com/Rosenkrands/sar-path-planning}} is used to determine the actions of the DCU for data acquisition and obstacle avoidance. If a static obstacle such as $\mathbb{R}_{\mathrm{NFZ}}$ or $\mathbb{R}_{\mathrm{BZ}}$ is encountered, the DCU performs detouring maneuvers. For dynamic obstacles like OUs, it either hovers in place or retreats to avoid collision until the OU moves away. To limit the noise impact when flying over $\mathbb{R}_{\mathrm{RZ}}$, the DCU must reduce its speed below $v_{\text{limit}}$ in accordance with regulatory requirements~\cite{6}. Furthermore, if the remaining energy is only sufficient for returning to $\mathbb{LA}$, the DCU will immediately head back.

  \item \textbf{LLM:} An LLM, DeepSeek-R1, is used in real-time to decide the actions of the DCU. We provide the LLM with the state information within a 20~m radius around the DCU, as well as the positions and remaining data of GEs. In addition, we also provide the LLM with $E_{\text{limit}}$ of the DCU.

    \item \textbf{SAC+Heuristic:} Similar to SAC+LLM, this variant employs the heuristic rule-based policy for obstacle avoidance instead of using the LLM. Specifically, when the DCU encounters potential collisions with OUs or static obstacles such as $\mathbb{R}_{\mathrm{NFZ}}$ or $\mathbb{R}_{\mathrm{BZ}}$, the heuristic controller overrides the SAC policy to execute avoidance or speed-reduction maneuvers according to predefined rules. This baseline serves to evaluate the effectiveness of integrating domain-specific expert knowledge into the SAC framework.

  \item \textbf{SAC+LLM (Fixed-$fp$):} SAC is used during training, with a certain probability ($fp \in \{0.3, 0.4, 0.5, 1.0\}$) of delegating action selection to the LLM instead of the policy network. The LLM used here is the same as the one employed in the proposed algorithm, i.e., DeepSeek-R1.

  \item \textbf{SAC:} A pure deep reinforcement learning approach using the SAC without LLM integration\footnote{We referenced the code from: \url{https://github.com/DLR-RM/stable-baselines3}}, which also implements PPO and DDPG, all available within the same repository.

  \item \textbf{PPO:} A representative on-policy reinforcement learning method, implemented within the same codebase as SAC, following the same architecture and parameterization.

  \item \textbf{DDPG:} A classic off-policy actor-critic method designed for continuous control, also implemented within the SAC codebase, ensuring consistency in experimental setup.

  \item \textbf{CPO:} A constrained reinforcement learning baseline based on safety‑starter‑agents from OpenAI\footnote{We referenced the code from: \url{https://github.com/openai/safety-starter-agents}}. It optimizes the expected cumulative reward while enforcing a safety constraint on the discounted cumulative cost. In our implementation, the instantaneous cost \(c_t\) is defined as a weighted combination of safety violations, including boundary collisions, OUs collisions, tall buildings collisions, and entries into NFZs. The threshold \(d\) is set to 0.05, corresponding to an acceptable violation rate of 5\%. The policy and value networks follow the same architecture as SAC, and the trust-region constraint on policy update is enforced with a maximum KL divergence of 0.01.

    \item \textbf{Existing methods:} We also include two recent RL approaches~\cite{theile2024equivariant, wang2023ensuring} for comparison to benchmark against state-of-the-art path planning and cooperative UAV control methods\footnote{We referenced the code for Theile et al.~\cite{theile2024equivariant} from: \url{https://github.com/theilem/uavSim}}%
    \footnote{We referenced the code for Wang et al.~\cite{wang2023ensuring} from: \url{https://github.com/BIT-MCS/DRL-UCS-AoI-Threshold}}.
\end{itemize}

\begin{figure}[!t]
\includegraphics[width=3.5in]{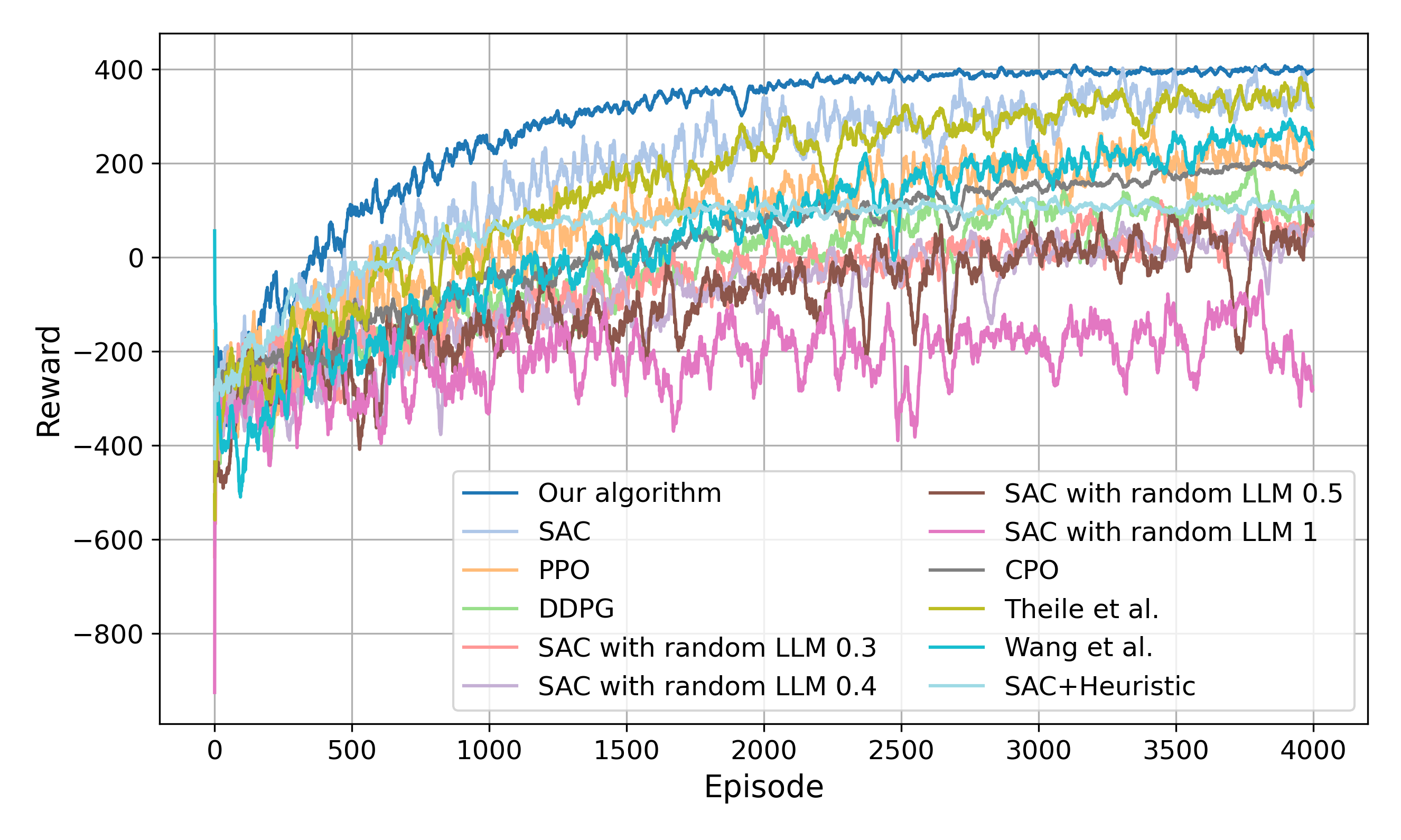}
\caption{Training reward curves of the proposed algorithm and all baseline methods over 4000 episodes, including SAC, PPO, DDPG, CPO, Theile et al.~\cite{theile2024equivariant}, Wang et al.~\cite{wang2023ensuring}, SAC+Heuristic, and SAC+LLM variants with different invocation probabilities.}
\label{fig3}
\vspace{-10pt}
\end{figure}

\subsubsection{Training Reward Curves about DRL Baselines and Our Algorithm}
We show the training reward curves of all DRL baselines and the proposed algorithm, as illustrated in Fig.~\ref{fig3}. Over 4000 training episodes, the proposed method achieves the highest and most stable cumulative rewards, indicating superior learning efficiency and policy quality. SAC and PPO converge well, with PPO showing slightly slower improvement, while DDPG reaches a lower reward plateau under complex constraints. The CPO algorithm exhibits stable but lower convergence due to its conservative optimization. Theile et al.~\cite{theile2024equivariant} achieves faster and more stable convergence through equivariant ensemble regularization, whereas Wang et al.~\cite{wang2023ensuring} converges slowly with lower final rewards. The SAC+Heuristic method shows early stability but limited long-term performance. Overall, the proposed hybrid LLM-augmented DRL algorithm achieves the best trade-off among efficiency, stability, and safety, outperforming both traditional and enhanced baselines.

\begin{figure*}[!t]
\centering
\includegraphics[width=6in]{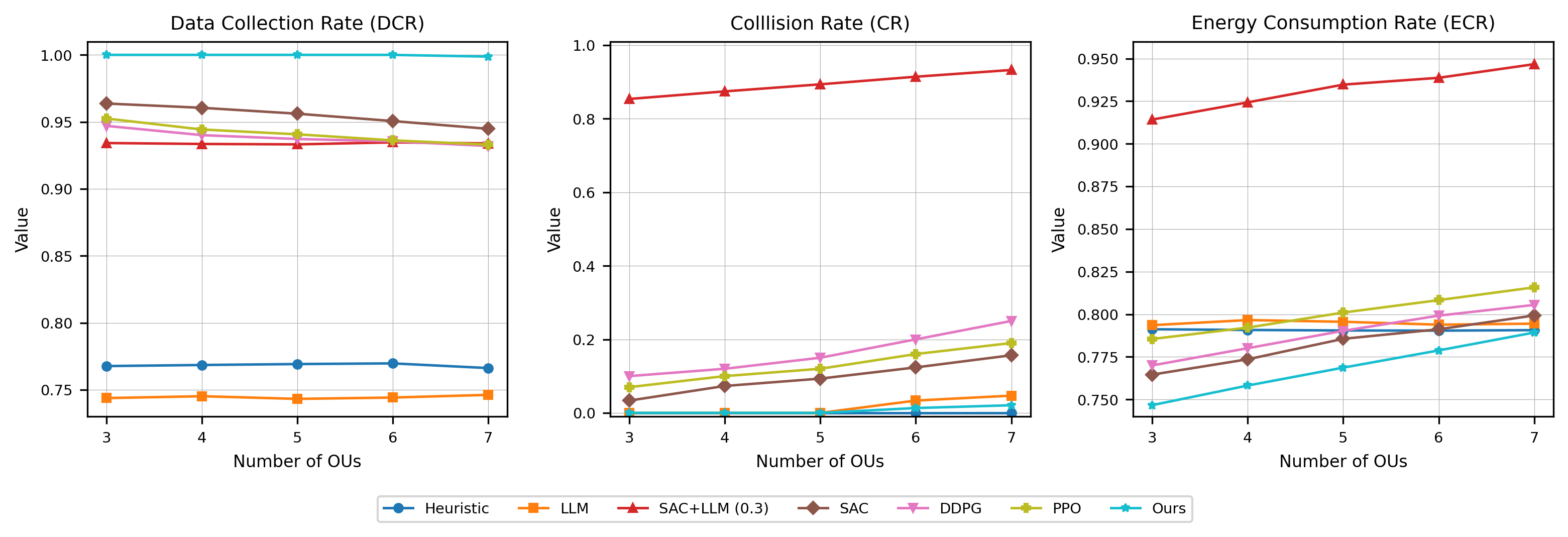}
\caption{The DCR, CR, and ECR about baselines and our algorithm when $N_{\mathrm{OU}}$ increases. The proposed algorithm consistently achieves the highest DCR with almost zero CR and lowest ECR, demonstrating superior efficiency and safety.}
\label{fig5}
\end{figure*}

\begin{figure*}[!t]
\centering
\includegraphics[width=6in]{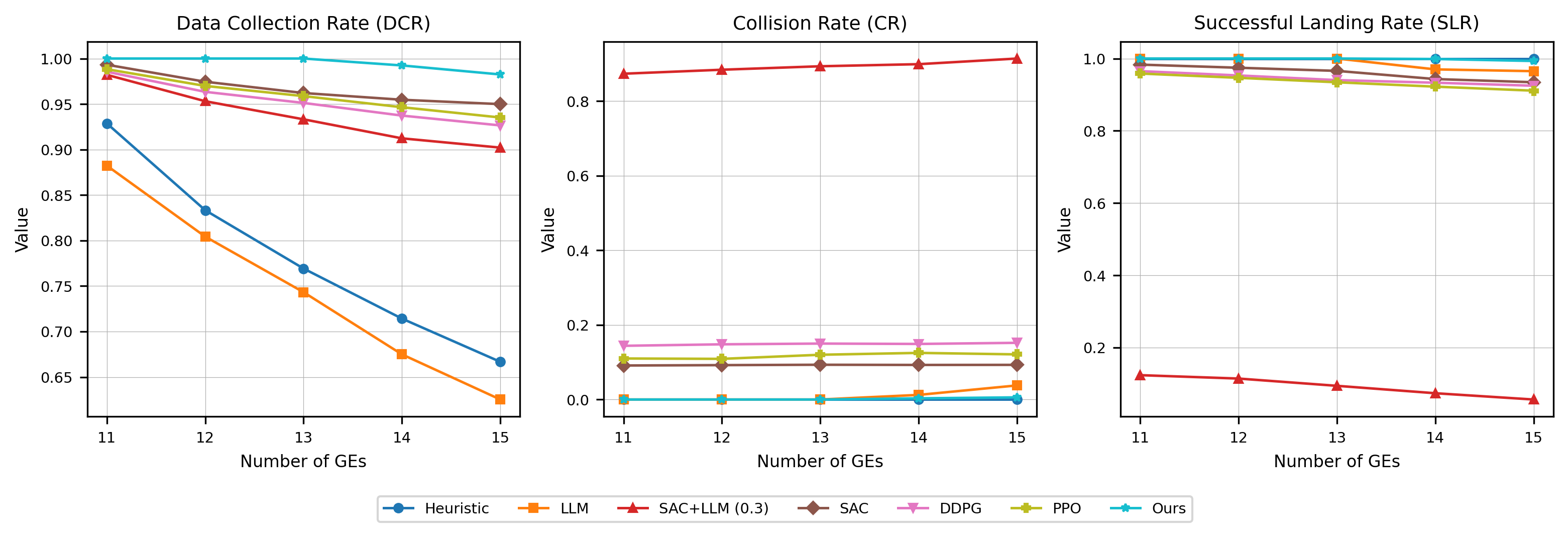}
\caption{The DCR, CR, and SLR about baselines and our algorithm when $N_{\mathrm{GE}}$ increases. Our algorithm maintains a stable and high DCR and CR of almost zero, while other baselines show declining performance and increased collisions. This highlights our algorithm’s robustness in complex environments.}
\label{fig6}
\end{figure*}

\begin{figure*}[!t]
\centering
\includegraphics[width=6in]{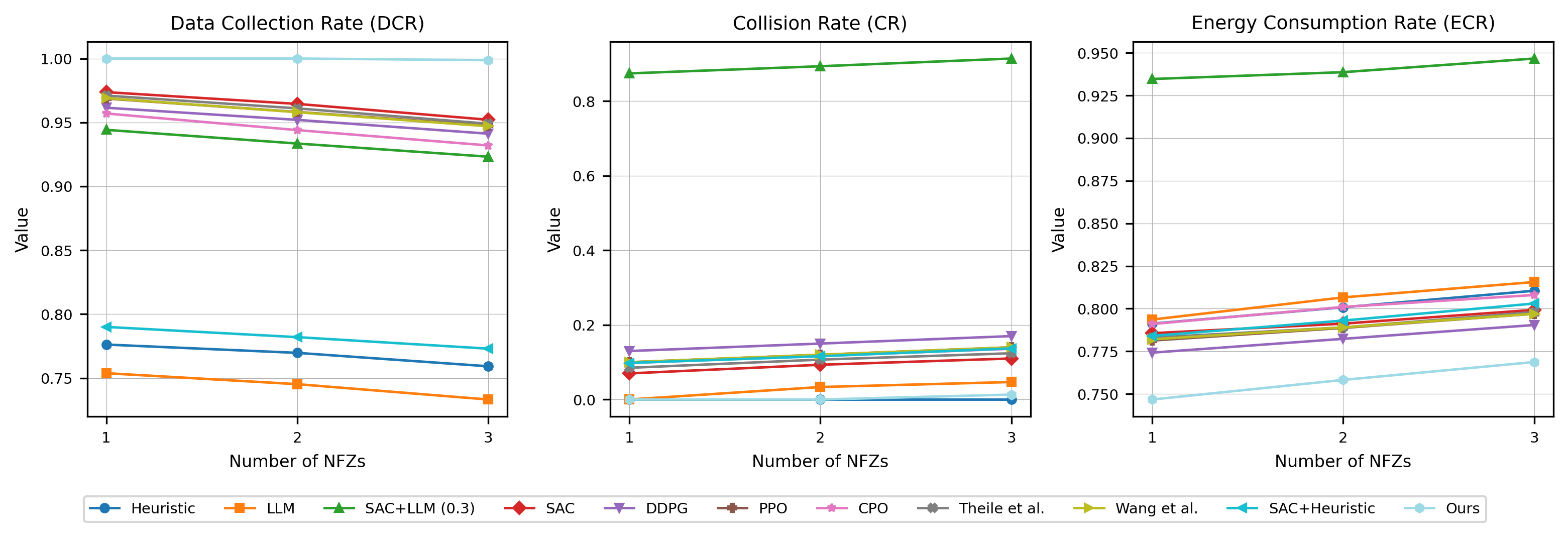}
\caption{The DCR, CR, and ECR about baselines and our algorithm when $N_{\mathrm{NFZ}}$ increases. While other algorithms suffer from reduced data collection or increased collisions, our algorithm sustains near-perfect DCR and near zero CR, with competitive energy consumption.}
\label{fig7}
\end{figure*}

\begin{figure*}[!t]
\centering
\includegraphics[width=6in]{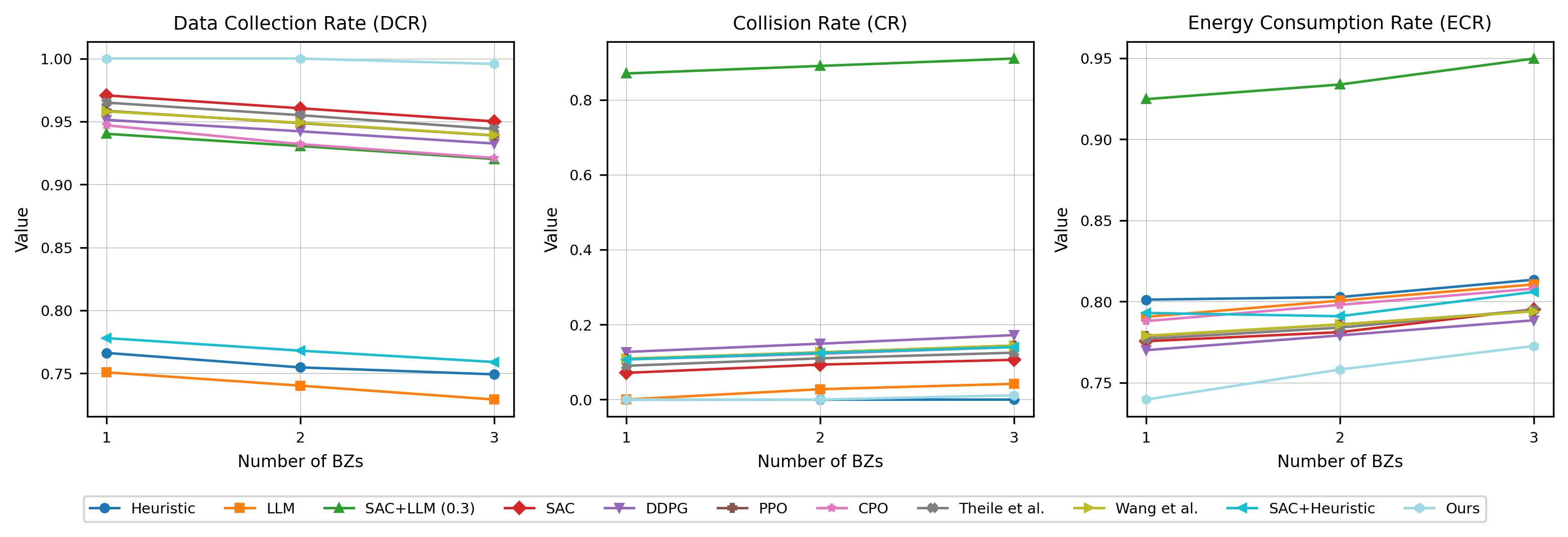}
\caption{The DCR, CR, and ECR about baselines and our algorithm when $N_{\mathrm{BZ}}$ increases. The proposed algorithm preserves a high DCR and almost avoids collisions entirely, while other baselines experience noticeable performance degradation.}
\label{fig8}
\end{figure*}

\begin{figure*}[!t]
\centering
\includegraphics[width=6in]{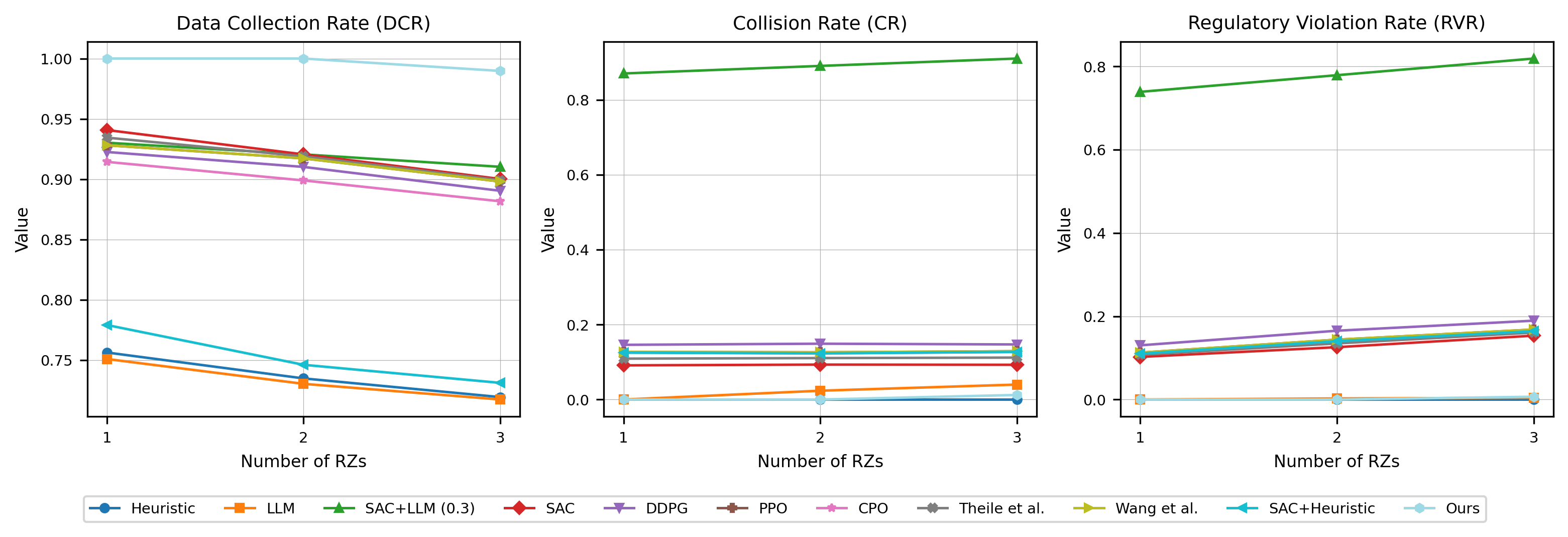}
\caption{The DCR, CR, and RVR about baselines and our algorithm when $N_{\mathrm{RZ}}$ increases. Our algorithm maintains full compliance (RVR 0), with high data collection efficiency and no collisions, while other baselines fails to adapt to regulatory constraints.}
\label{fig9}
\end{figure*}

\begin{figure}[!t]
\includegraphics[width=3in]{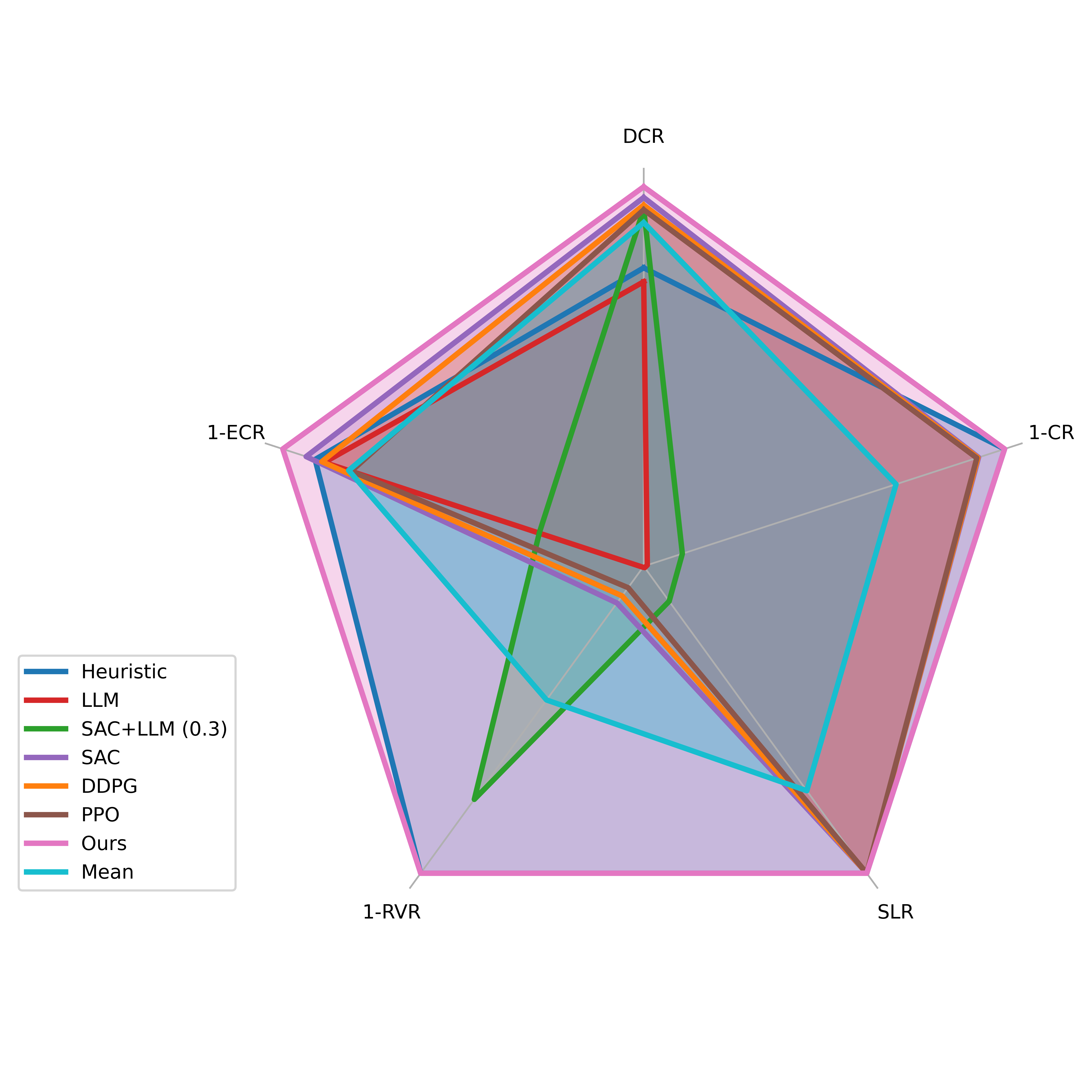}
\caption{Average performance summary across all environmental configurations. Our algorithm achieves the highest DCR (99.49\%), lowest CR (0\%), and strong compliance (RVR 0\% and ECR 76.95\%), validating its overall effectiveness across diverse low-altitude scenarios.}
\label{fig10}
\vspace{-10pt}
\end{figure}

\begin{figure*}[!t]
\centering
\includegraphics[width=6.5in]{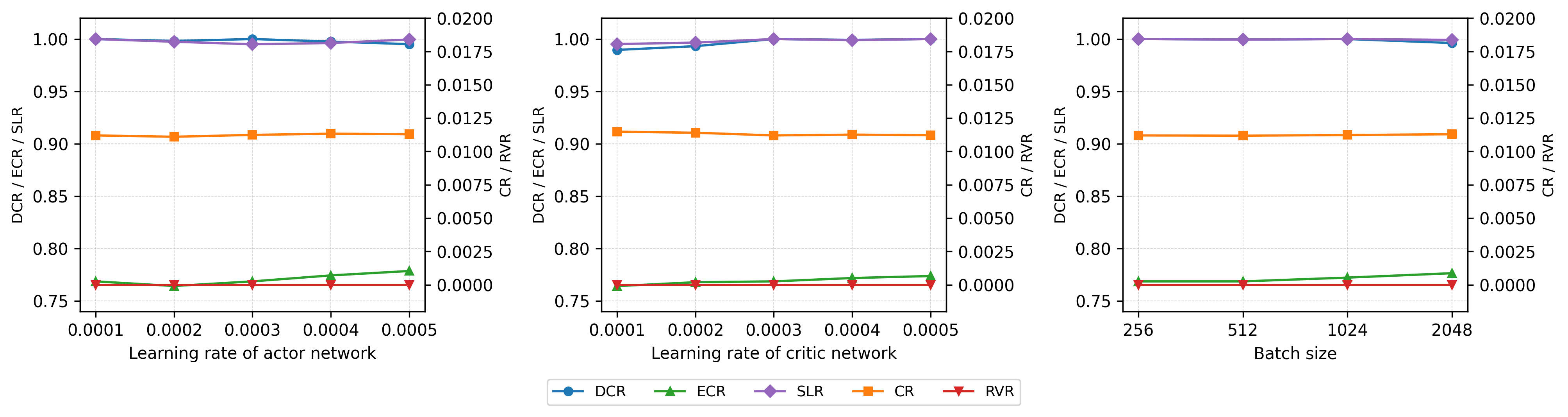}
\caption{Variations of DCR, CR, ECR, SLR, and RVR under different hyperparameter settings of (a) actor LR, (b) critic LR, and (c) batch size.}
\label{fig11}
\vspace{-10pt}
\end{figure*}

\begin{figure*}[!t]
\centering
\includegraphics[width=7in]{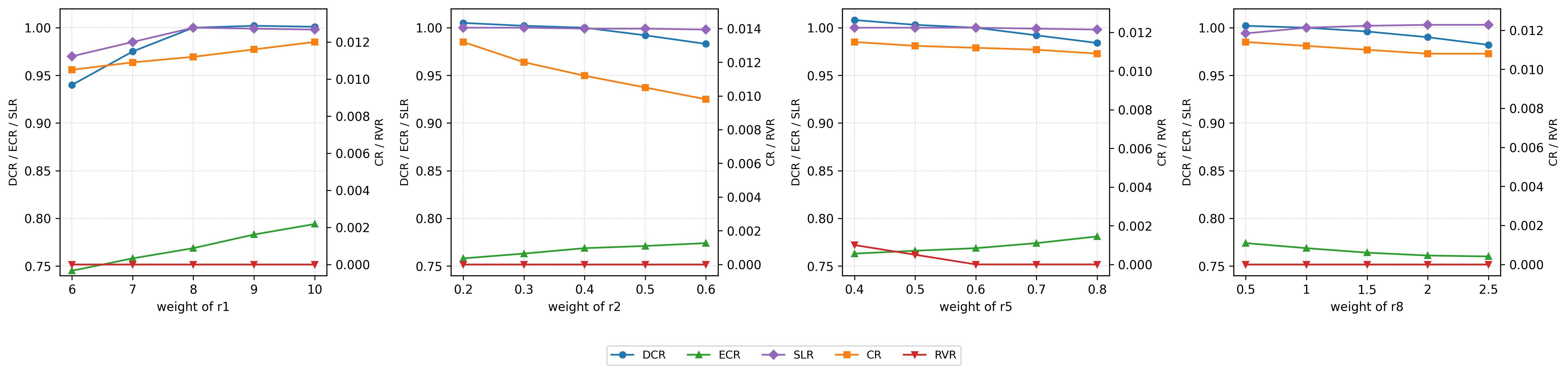}
\caption{Variations of DCR, CR, ECR, SLR, and RVR under different reward weight settings for $r_1$, $r_2$, $r_5$, and $r_8$.}
\label{fig12}
\vspace{-10pt}
\end{figure*}

\begin{figure}[t]
\centering
\includegraphics[width=0.35\textwidth]{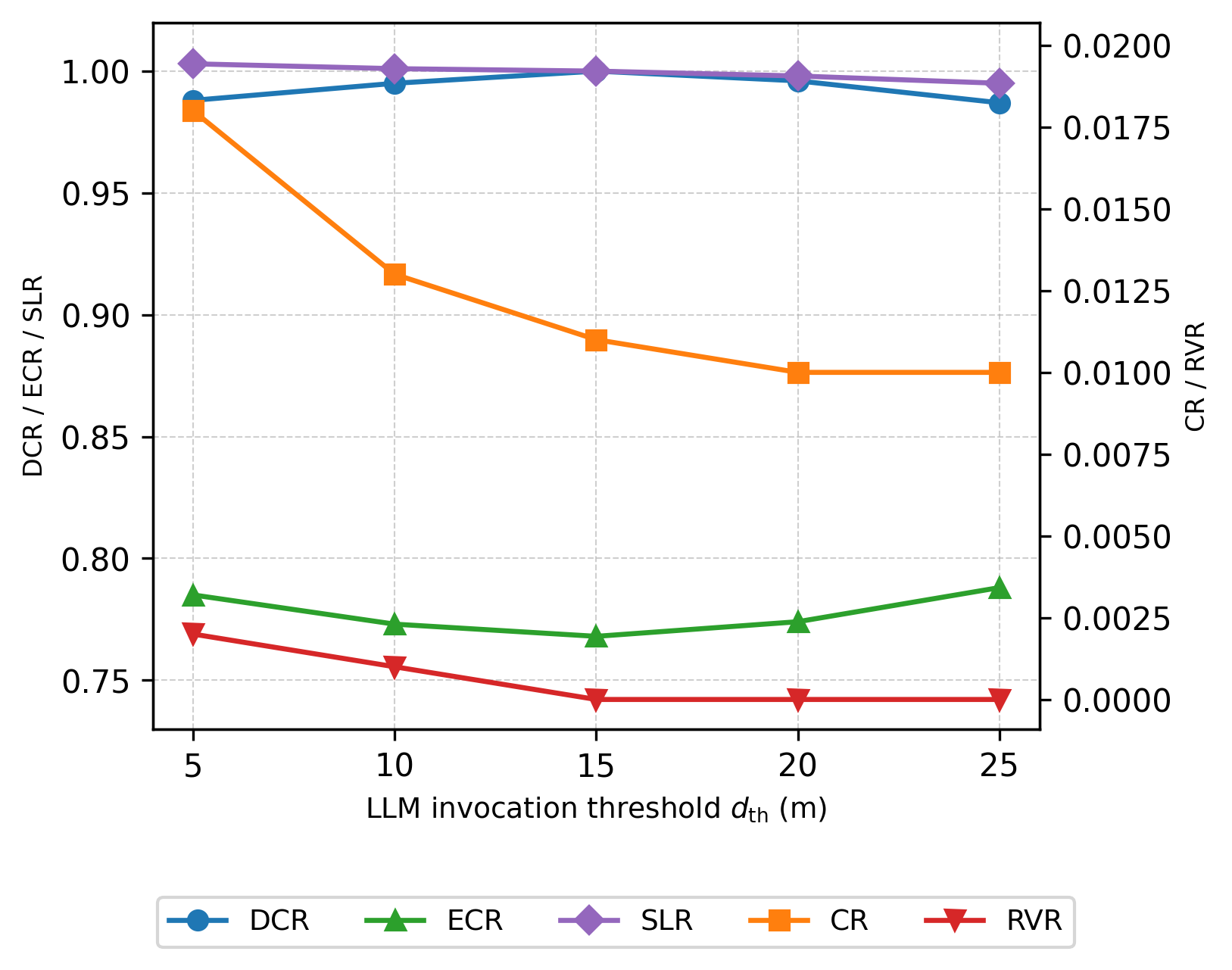}
\caption{Sensitivity analysis of the LLM invocation threshold the threshold. 
The figure shows the impact of varying $d_{\text{th}} \in \{5,10,15,20,25\}$~m on five key performance metrics (DCR, CR, ECR, SLR, and RVR). 
Excessively small thresholds (e.g., 5~m) cause delayed LLM activation, leading to higher collision rates and lower DCR, while overly large thresholds (e.g., 25~m) result in frequent LLM invocations and increased energy consumption. 
The threshold of 15~m achieves the best trade-off between safety, efficiency, and energy consumption, validating the choice used in the main experiments.}
\label{fig13}
\end{figure}

\begin{figure}[t]
  \centering
  \includegraphics[width=\linewidth]{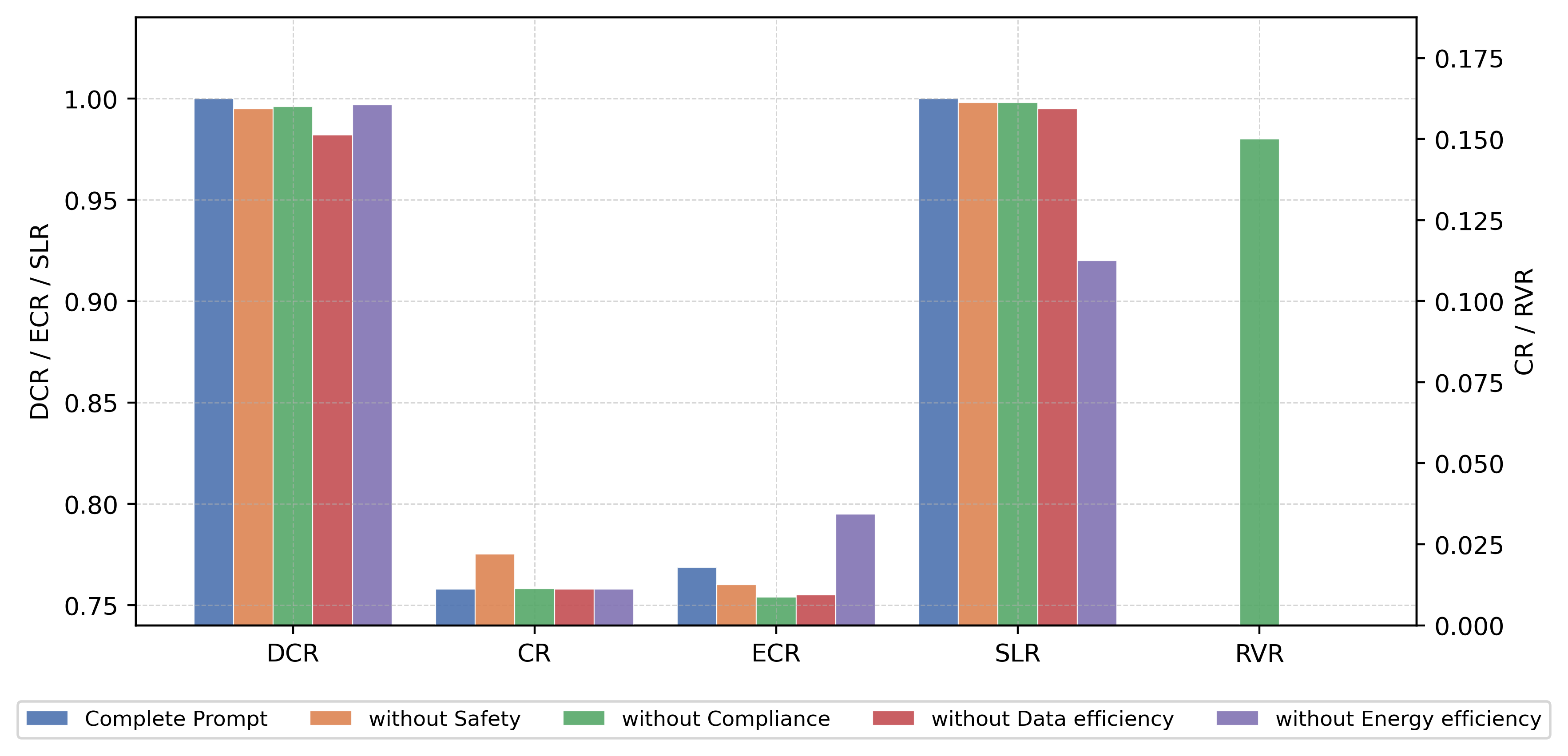}
  \caption{Ablation on reasoning objectives embedded in the LLM prompt. 
  Bars compare the \emph{Complete Prompt} with four variants that remove one objective at a time 
  (\emph{without Safety}, \emph{without Compliance}, \emph{without Data efficiency}, \emph{without Energy efficiency}) 
  across five metrics (DCR, CR, ECR, SLR, RVR).}
  \label{fig:prompt_ablation}
\end{figure}

\subsubsection{Metrics about Baselines and Our Algorithm}
The experimental results in Figs.~5--10 evaluate five core metrics (DCR, CR, SLR, RVR, and ECR) under varying numbers of OUs, GEs, NFZs, BZs, and RZs, verifying how our algorithm satisfies the requirements of robust obstacle avoidance (\textbf{R2}), compliance (\textbf{R3}), and economy (\textbf{R4}) while addressing \textbf{Challenges~1--3}.

Fig.~\ref{fig5} shows the performance under increasing OUs. Our method consistently achieves the highest DCR and the lowest CR and ECR, confirming its robustness and energy efficiency in dynamic environments. SAC and PPO converge well but suffer from rising CR, while SAC+LLM becomes unstable with CR exceeding 90\%. DDPG performs moderately, and the standalone LLM and heuristic methods remain conservative with limited DCR. Among new baselines, CPO achieves slightly lower DCR but lower CR than PPO, Theile et al.~\cite{theile2024equivariant} and Wang et al.~\cite{wang2023ensuring} perform between SAC and PPO, and SAC+Heuristic outperforms the heuristic baseline with better safety and efficiency.

Fig.~\ref{fig6} presents the results under increasing GEs. Our algorithm maintains DCR $\approx$ 1.0 and SLR $\approx$ 1.0, showing excellent reliability. SAC, PPO, and DDPG exhibit moderate declines in DCR and SLR as complexity rises. SAC+LLM collapses under high complexity, while CPO reduces CR at the cost of minor DCR loss. Theile et al.~\cite{theile2024equivariant} and Wang et al.~\cite{wang2023ensuring} remain close to PPO, and SAC+Heuristic balances performance and safety better than the rule-based baseline. These results demonstrate the superior adaptability of our method to complex mission scenarios.

Figs.~\ref{fig7} and~\ref{fig8} examine static obstacle cases (NFZs and BZs). Our approach maintains DCR $\approx$ 1.0 and CR $\approx$ 0, achieving the best obstacle avoidance and lowest ECR. SAC and PPO perform well but show rising CR with denser obstacles, while CPO achieves slightly lower DCR yet improved safety. Theile et al.~\cite{theile2024equivariant} and Wang et al.~\cite{wang2023ensuring} remain moderate, and SAC+Heuristic improves both DCR and CR over the heuristic baseline. These confirm the algorithm’s effectiveness in static, constrained environments (\textbf{Challenges~1--2}).

Fig.~\ref{fig9} evaluates compliance under increasing RZs. Our method sustains the highest DCR with minimal CR and RVR, indicating excellent trade-offs between efficiency and regulatory adherence. SAC+LLM achieves high DCR but violates more frequently, while SAC, PPO, and DDPG show moderate compliance. CPO further lowers violations, Theile et al.~\cite{theile2024equivariant} and Wang et al.~\cite{wang2023ensuring} stay between SAC and PPO, and SAC+Heuristic enhances both safety and DCR compared to heuristic control. Overall, our algorithm uniquely balances safety, compliance, and efficiency under soft constraints (\textbf{Challenge~3}).

The radar chart in Fig.~\ref{fig10} clearly demonstrates the superior performance of our proposed algorithm across all five metrics. It achieves high DCR and low ECR, indicating effective data collection and energy-efficient mission completion (\textbf{R4}). Simultaneously, it maintains high SLR and low CR and RVR, ensuring both obstacle avoidance and regulatory compliance, critical for safe operation in urban, airspace-reused environments (\textbf{R2} and \textbf{R3}).

In summary, the results validate that our algorithm, by selectively engaging the LLM for obstacle avoidance, empowers the DCU to make informed and adaptive decisions in complex, dynamic environments. This targeted integration enables our algorithm to uniquely balance the three core demands of the low-altitude economy: \textbf{R2}, \textbf{R3}, and \textbf{R4}. By meeting these requirements, our algorithm effectively addresses the corresponding \textbf{Challenges~1--3}. No other baseline achieves comparable comprehensive performance across these dimensions. Thus, our algorithm stands out as a practical solution for urban low-altitude data collection trajectory planning scenarios.

\subsubsection{Parameters Sensitive Analysis}
To comprehensively evaluate the robustness and generalization capability of the proposed algorithm, we conduct a series of sensitivity analyses covering four aspects: 
(1) hyperparameters of the training process, 
(2) weight coefficients of the reward function, 
(3) the invocation threshold distance for triggering the LLM-based decision process, and 
(4) the reasoning objectives embedded in the LLM prompt. 
The detailed analyses and corresponding results are presented as follows.

As shown in Fig.~\ref{fig11}, variations in the actor and critic learning rates and batch size cause only minor fluctuations in performance. 
DCR and SLR stay close to~1.0, while CR and RVR remain nearly zero, indicating stable learning and strong safety. 
ECR varies slightly within~[0.76, 0.78], showing consistent energy efficiency. 
Overall, the proposed model exhibits high robustness to hyperparameter changes.

To evaluate the impact of reward composition, we vary the weights of $r_1$, $r_2$, $r_5$, and $r_8$, which correspond to task efficiency, obstacle avoidance, compliance, and energy return, respectively. As shown in Fig.~\ref{fig12}, increasing $r_1$ enhances DCR and SLR but slightly raises ECR and CR, while higher $r_2$ effectively lowers CR at a small cost of DCR. Increasing $r_5$ improves compliance (lower RVR) but slightly reduces efficiency, and larger $r_8$ values yield better energy use and landing stability with a minor DCR drop. Overall, the proposed algorithm remains stable and well-balanced under moderate reward weight variations.

As shown in Fig.~\ref{fig13}, the variation of the LLM invocation threshold exhibits a clear trade-off among safety, efficiency, and energy consumption. 
When the threshold is too small (e.g., 5~m), the LLM is triggered too late, leading to occasional collisions and reduced data collection efficiency. 
Conversely, excessively large thresholds (e.g., 25~m) cause premature or frequent invocations, increasing energy consumption and slightly reducing stability. 
Moderate thresholds such as 10~m and 20~m achieve similar performance to 15~m, but the latter offers the best balance, maintaining high DCR and SLR with low CR and ECR. 
Therefore, $15$~m is adopted as the default setting in all subsequent experiments.

We ablate the four reasoning objectives in the LLM prompt by removing one component at a time and compare against the complete prompt shown in Fig.~\ref{fig:cot_prompt_generator}. The result is shown in Fig.~\ref{fig:prompt_ablation}.
Removing \emph{Safety} increases CR and slightly reduces DCR, confirming its primary role in risk control.
Removing \emph{Compliance} markedly degrades RVR despite similar CR, indicating the necessity of explicit compliance guidance.
Removing \emph{Data efficiency} lowers DCR and slightly changes ECR, showing that prioritizing informative targets contributes to sustained task completion.
Removing \emph{Energy efficiency} harms energy behavior and landing stability, demonstrating that explicit energy-aware reasoning helps maintain safe mission termination.
Overall, the complete prompt provides the best balance across safety, compliance, and efficiency.

In summary, all sensitivity analyses consistently demonstrate the robustness and adaptability of the proposed algorithm under diverse parameter and configuration settings. 
Neither hyperparameter tuning nor moderate reward reweighting leads to noticeable performance degradation, and both the LLM invocation threshold and reasoning objectives exhibit stable behavior with interpretable effects. 
These results collectively validate that the proposed hybrid SAC–LLM algorithm maintains a well-balanced trade-off among safety, compliance, efficiency, and energy performance, even when the underlying parameters or decision conditions vary across different environments.

\section{Conclusion}

This paper developed a hybrid unmanned aerial vehicle (UAV) trajectory planning algorithm that combines deep reinforcement learning with large language model reasoning for low-altitude data collection in complex urban environments. The proposed algorithm jointly considers obstacle avoidance, regulation awareness, and energy efficiency, achieving safe and adaptive decision-making under uncertainty.

Future work will explore extending the algorithm to multi-UAV collaboration and large-scale city scenarios. 
Another direction is to enhance three-dimensional obstacle avoidance and communication-aware coordination under partial or unstable network conditions. 
In addition, incorporating more realistic air-to-ground channel models, including probabilistic line-of-sight and non-line-of-sight propagation in urban environments, will help capture complex communication dynamics. 
Furthermore, integrating real sensor noise, multimodal perception, and dynamic regulatory feedback will be essential for improving robustness and real-world applicability in low-altitude economy networks.


\begin{thebibliography}{99}

\bibitem{15}
R. Zhang, J. He, X. Luo, D. Niyato, J. Kang, Z. Xiong, Y. Li, and B. Sikdar, ``Toward democratized generative AI in next‑generation mobile edge networks,'' \emph{IEEE Network}, early access, 2025, doi: 10.1109/MNET.2025.3541078.

\bibitem{16}
R. Zhang, H. Du, Y. Liu, D. Niyato, J. Kang, Z. Xiong, A. Jamalipour, and D. I. Kim, ``Generative AI agents with large language model for satellite networks via a mixture of experts transmission,'' \emph{IEEE Journal on Selected Areas in Communications}, vol. 42, no. 12, pp. 3581--3596, Dec. 2024, doi: 10.1109/JSAC.2024.3459037.



\bibitem{2}
M. Ahmed, A. A. Soofi, F. Khan, S. Raza, W. U. Khan, L. Su, F. Xu, and Z. Han, ``Toward a sustainable low-altitude economy: A survey of energy-efficient RIS-UAV networks,'' \emph{arXiv preprint}, arXiv:2504.02162, Apr. 2025. [Online]. Available: \url{https://arxiv.org/abs/2504.02162}.

\bibitem{18}
R. Zhang, K. Xiong, Y. Lu, P. Fan, D. W. K. Ng, and K. B. Letaief, ``Energy efficiency maximization in RIS-assisted SWIPT networks with RSMA: A PPO-based approach,'' \emph{IEEE Journal on Selected Areas in Communications}, vol. 41, no. 5, pp. 1413--1430, May 2023, doi: 10.1109/JSAC.2023.3240707.

\bibitem{17}
R. Zhang \emph{et al.}, ``Generative AI for Space-Air-Ground Integrated Networks,'' \emph{IEEE Wireless Communications}, vol. 31, no. 6, pp. 10--20, Dec. 2024, doi: 10.1109/MWC.016.2300547.


\bibitem{5}
N. Zhang, M. Zhang, and K. H. Low, ``3D trajectory planning and real-time collision resolution of multirotor drone operations in complex urban low-altitude airspace,'' \emph{Transportation Research Part C: Emerging Technologies}, vol. 129, p. 103123, 2021, doi: 10.1016/j.trc.2021.103123.

\bibitem{6}
H. J. Hadi, Y. Cao, K. U. Nisa, Y. Mekdad, A. Aris, L. Babun, A. E. Fergougui, M. Conti, R. Lazzeretti, and A. S. Uluagac, ``A comprehensive survey on security, privacy issues and emerging defence technologies for UAVs,'' \emph{Journal of Network and Computer Applications}, vol. 213, p. 103607, 2023, doi: 10.1016/j.jnca.2023.103607.

\bibitem{7}
X. Qin, Z. Song, T. Hou, W. Yu, J. Wang, and X. Sun, ``Joint optimization of resource allocation, phase shift, and UAV trajectory for energy-efficient RIS-assisted UAV-enabled MEC systems,'' \emph{IEEE Transactions on Green Communications and Networking}, vol. 7, no. 4, pp. 1778--1792, Dec. 2023, doi: 10.1109/TGCN.2023.3287604.

\bibitem{8}
H. Pan, Y. Liu, G. Sun, J. Fan, S. Liang, and C. Yuen, ``Joint power and 3D trajectory optimization for UAV-enabled wireless powered communication networks with obstacles,'' \emph{IEEE Transactions on Communications}, vol. 71, no. 4, pp. 2364--2380, Apr. 2023, doi: 10.1109/TCOMM.2023.3240697.

\bibitem{9}
F. Pervez, A. Sultana, C. Yang, and L. Zhao, ``Energy and latency efficient joint communication and computation optimization in a multi-UAV-assisted MEC network,'' \emph{IEEE Transactions on Wireless Communications}, vol. 23, no. 3, pp. 1728--1741, Mar. 2024, doi: 10.1109/TWC.2023.3291692.

\bibitem{10}
Y. Zhang, Y. Huang, C. Huang, H. Huang, and A.-T. Nguyen, ``Joint optimization of deployment and flight planning of multi-UAVs for long-distance data collection from large-scale IoT devices,'' \emph{IEEE Internet of Things Journal}, vol. 11, no. 1, pp. 791--804, Jan. 2024, doi: 10.1109/JIOT.2023.3285942.

\bibitem{11}
K. Heo, G. Park, and K. Lee, ``Joint optimization of UAV trajectory and communication resources with complete avoidance of no-fly-zones,'' \emph{IEEE Transactions on Intelligent Transportation Systems}, vol. 25, no. 10, pp. 14259--14265, Oct. 2024, doi: 10.1109/TITS.2024.3403887.

\bibitem{12}
J. Li, G. Sun, L. Duan, and Q. Wu, ``Multi-objective optimization for UAV swarm-assisted IoT with virtual antenna arrays,'' \emph{IEEE Transactions on Mobile Computing}, vol. 23, no. 5, pp. 4890--4907, May 2024, doi: 10.1109/TMC.2023.3298888.

\bibitem{13}
Z. Fu, J. Liu, Y. Mao, L. Q. Qu, L. F. Xie, and X. Wang, ``Energy-efficient UAV-assisted federated learning: Trajectory optimization, device scheduling, and resource management,'' \emph{IEEE Transactions on Network and Service Management}, vol. 22, no. 2, pp. 974--988, Jun. 2025, doi: 10.1109/TNSM.2025.3531237.

\bibitem{14}
Z. Wang, J. Wen, J. He, L. Yu, and Z. Li, ``Energy efficiency optimization of RIS-assisted UAV search-based cognitive communication in complex obstacle avoidance environments,'' \emph{IEEE Transactions on Cognitive Communications and Networking}, early access, Feb. 2025, doi: 10.1109/TCCN.2025.3544267.


\bibitem{19}
S. Silvirianti, B. N. Narottama, and S. Y. Shin, ``Layerwise quantum deep reinforcement learning for joint optimization of UAV trajectory and resource allocation,'' \emph{IEEE Internet of Things Journal}, vol. 11, no. 1, pp. 430--443, Jan. 2024, doi: 10.1109/JIOT.2023.3285968.

\bibitem{20}
Y. Chen, Y. Yang, Y. Wu, J. Huang, and L. Zhao, ``Joint trajectory optimization and resource allocation in UAV-MEC systems: A Lyapunov-assisted DRL approach,'' \emph{IEEE Transactions on Services Computing}, early access, Feb. 2025, doi: 10.1109/TSC.2025.3544124.

\bibitem{21}
Z. Ning, Y. Yang, X. Wang, Q. Song, L. Guo, and A. Jamalipour, ``Multi-agent deep reinforcement learning based UAV trajectory optimization for differentiated services,'' \emph{IEEE Transactions on Mobile Computing}, vol. 23, no. 5, pp. 5818--5834, May 2024, doi: 10.1109/TMC.2023.3312276.

\bibitem{22}
F. Song, M. Deng, H. Xing, Y. Liu, F. Ye, and Z. Xiao, ``Energy-efficient trajectory optimization with wireless charging in UAV-assisted MEC based on multi-objective reinforcement learning,'' \emph{IEEE Transactions on Mobile Computing}, vol. 23, no. 12, pp. 10867--10884, Dec. 2024, doi: 10.1109/TMC.2024.3384405.

\bibitem{23}
R. Ding, F. Zhou, Q. Wu, and D. W. K. Ng, ``From external interaction to internal inference: An intelligent learning framework for spectrum sharing and UAV trajectory optimization,'' \emph{IEEE Transactions on Wireless Communications}, vol. 23, no. 9, pp. 12099--12114, Sept. 2024, doi: 10.1109/TWC.2024.3387980.

\bibitem{24}
T. Wang, W. Du, C. Jiang, Y. Li, and H. Zhang, ``Safety constrained trajectory optimization for completion time minimization for UAV communications,'' \emph{IEEE Internet of Things Journal}, vol. 11, no. 21, pp. 34482--34491, Nov. 2024, doi: 10.1109/JIOT.2024.3355906.

\bibitem{25}
H. He, W. Yuan, S. Chen, X. Jiang, F. Yang, and J. Yang, ``Deep reinforcement learning-based distributed 3D UAV trajectory design,'' \emph{IEEE Transactions on Communications}, vol. 72, no. 6, pp. 3736--3751, Jun. 2024, doi: 10.1109/TCOMM.2024.3361534.

\bibitem{26}
Z. Liu, J. Zhang, Y. Zeng, and B. Ai, ``Energy-efficient multi-agent reinforcement learning for UAV trajectory optimization in cell-free massive MIMO networks,'' \emph{IEEE Transactions on Wireless Communications}, early access, Mar. 2025, doi: 10.1109/TWC.2025.3550266.

\bibitem{27}
Z. Ning, H. Ji, X. Wang, E. C. H. Ngai, L. Guo, and J. Liu, ``Joint optimization of data acquisition and trajectory planning for UAV-assisted wireless powered Internet of Things,'' \emph{IEEE Transactions on Mobile Computing}, vol. 24, no. 2, pp. 1016--1030, Feb. 2025, doi: 10.1109/TMC.2024.3470831.

\bibitem{28}
J. Zhong, M. Li, Y. Chen, Z. Wei, F. Yang, and H. Shen, ``A safer vision-based autonomous planning system for quadrotor UAVs with dynamic obstacle trajectory prediction and its application with LLMs,'' in \emph{Proc. IEEE/CVF Winter Conference on Applications of Computer Vision Workshops (WACV Workshops)}, 2024, pp. 920--929.

\bibitem{29}
A. Phadke, A. Hadimlioglu, T. Chu, and C. N. Sekharan, ``Integrating large language models for UAV control in simulated environments: A modular interaction approach,'' \emph{arXiv preprint}, arXiv:2410.17602, 2024. [Online]. Available: \url{https://arxiv.org/abs/2410.17602}.

\bibitem{30}
J. Xiao, C. Tsao, Y. Zhang, and M. Feroskhan, ``FM-Planner: Foundation model guided trajectory planning for autonomous drone navigation,'' \emph{arXiv preprint}, arXiv:2505.20783, 2025. [Online]. Available: \url{https://arxiv.org/abs/2505.20783}.

\bibitem{31}
H. Samma and S. El-Ferik, ``UAV visual trajectory planning using large language models,'' \emph{Transportation Research Procedia}, vol. 84, pp. 339--345, 2025, doi: 10.1016/j.trpro.2025.03.081.

\bibitem{32}
S. Cai, Y. Wu, and L. Zhou, ``LLM-Land: Large language models for context-aware drone landing,'' \emph{arXiv preprint}, arXiv:2505.06399, 2025. [Online]. Available: \url{https://arxiv.org/abs/2505.06399}.

\bibitem{33}
P. Li, Z. An, S. Abrar, and L. Zhou, ``Large language models for multi-robot systems: a survey,'' \emph{arXiv preprint}, arXiv:2502.03814, Feb. 2025. [Online]. Available: \url{https://arxiv.org/abs/2502.03814}.

\bibitem{34}
P. Razzaghi \emph{et al.}, ``A survey on reinforcement learning in aviation applications,'' \emph{Engineering Applications of Artificial Intelligence}, vol. 115, pp. 1--18, Aug. 2023.


\bibitem{36}
Y. Wang, Z. Wei, H. Wu, and Z. Feng, ``Toward Realization of Low-Altitude Economy Networks: Core Architecture, Integrated Technologies, and Future Directions,'' \emph{arXiv preprint}, arXiv:2504.21583, Apr. 2025. [Online]. Available: \url{https://arxiv.org/abs/2504.21583}.

\bibitem{37}
Z. Wei \emph{et al.}, ``UAV-assisted data collection for Internet of Things: A survey,'' \emph{IEEE Internet of Things Journal}, vol. 9, no. 17, pp. 15460--15483, Sept. 2022, doi: 10.1109/JIOT.2022.3182483.

\bibitem{38}
D. D. Falconer, F. Adachi, and B. Gudmundson, ``Time division multiple access methods for wireless personal communications,'' \emph{IEEE Communications Magazine}, vol. 33, no. 1, pp. 50--57, Jan. 1995.

\bibitem{39}
Y. Li, ``Deep reinforcement learning: An overview,'' \emph{arXiv preprint}, arXiv:1701.07274, Jan. 2017. [Online]. Available: \url{https://arxiv.org/abs/1701.07274}.

\bibitem{40}
X. Tang, Y. Wang, Q. Huang, Y. Li, and C. Wang, ``Highway decision-making and motion planning for autonomous driving via soft actor-critic,'' \emph{IEEE Transactions on Vehicular Technology}, vol. 71, no. 5, pp. 4706--4717, May 2022, doi: 10.1109/TVT.2022.3151651.

\bibitem{41}
Q. Zhang \emph{et al.}, ``Data-Aided Doppler Frequency Shift Estimation and Compensation for UAVs,'' \emph{IEEE Internet of Things Journal}, vol. 7, no. 1, pp. 400--415, Jan. 2020, doi: 10.1109/JIOT.2019.2943608.

\bibitem{42}
H.-T. Ye, X. Kang, J. Joung, and Y.-C. Liang, ``Optimization for Full-Duplex Rotary-Wing UAV-Enabled Wireless-Powered IoT Networks,'' \emph{IEEE Transactions on Wireless Communications}, vol. 19, no. 7, pp. 5057--5072, Jul. 2020, doi: 10.1109/TWC.2020.2989302.

\bibitem{44}
C. Wen, L. Qiu, and X. Liang, ``Securing UAV communication with mobile UAV eavesdroppers: Joint trajectory and communication design,'' in \emph{Proc. IEEE Wireless Communications and Networking Conference (WCNC)}, 2021, pp. 1--6, doi: 10.1109/WCNC49053.2021.9417318.

\bibitem{45}
X. Wang, M. C. Gursoy, T. Erpek, and Y. E. Sagduyu, ``Collision-aware UAV trajectories for data collection via reinforcement learning,'' in \emph{Proc. IEEE Global Communications Conference (GLOBECOM)}, 2021, pp. 1--6, doi: 10.1109/GLOBECOM46510.2021.9686015.

\bibitem{46}
S. Rahim, L. Peng, and P.-H. Ho, ``TinyFDRL-Enhanced Energy-Efficient Trajectory Design for Integrated Space-Air-Ground Networks,'' \emph{IEEE Internet of Things Journal}, pp. 1--1, 2024, doi: 10.1109/JIOT.2024.3361394.

\bibitem{47}
Y. Zeng, J. Xu, and R. Zhang, ``Energy Minimization for Wireless Communication With Rotary-Wing UAV,'' \emph{IEEE Transactions on Wireless Communications}, vol. 18, no. 4, pp. 2329--2345, Apr. 2019, doi: 10.1109/TWC.2019.2902559.

\bibitem{48}
D. Chen and W. Hua, ``Hierarchical VAE Based Semantic Communications for POMDP Tasks,'' in \emph{Proc. IEEE Int. Conf. Acoust., Speech Signal Process. (ICASSP)}, Apr. 2024, pp. 5540--5544, doi: 10.1109/ICASSP48485.2024.10445833.

\bibitem{49}
DeepSeek, ``DeepSeek: Large Language Models and Multimodal Intelligence,'' [Online]. Available: \url{https://www.deepseek.com/}. [Accessed: May 27, 2025].

\bibitem{50}
X. Wang, M. C. Gursoy, T. Erpek, and Y. E. Sagduyu, ``Learning-Based UAV Trajectory planning for Data Collection With Integrated Collision Avoidance,'' \emph{IEEE Internet of Things Journal}, vol. 9, no. 17, pp. 16663--16676, Sep. 2022, doi: 10.1109/JIOT.2022.3153585.

\bibitem{51}
J. Fan \emph{et al.}, ``Energy-constrained safe trajectory planning for UAV-assisted data collection of mobile IoT devices,'' \emph{IEEE Internet of Things Journal}, vol. 11, no. 24, pp. 39971--39983, Dec. 2024, doi: 10.1109/JIOT.2024.3448537.

\bibitem{52}
T. Haarnoja, A. Zhou, P. Abbeel, and S. Levine, ``Soft Actor-Critic: Off-Policy Maximum Entropy Deep Reinforcement Learning with a Stochastic Actor,'' in \emph{Proc. 35th Int. Conf. Mach. Learn. (ICML)}, Stockholm, Sweden, 2018, pp. 1861--1870.


\bibitem{54}
J. Kaplan \emph{et al.}, ``Scaling Laws for Neural Language Models,'' \emph{arXiv preprint}, arXiv:2001.08361, 2020. [Online]. Available: \url{https://arxiv.org/abs/2001.08361}.

\bibitem{55}
J. Schulman, F. Wolski, P. Dhariwal, A. Radford, and O. Klimov, ``Proximal Policy Optimization Algorithms,'' \emph{arXiv preprint}, arXiv:1707.06347, Jul. 2017. [Online]. Available: \url{https://arxiv.org/abs/1707.06347}.

\bibitem{56}
T. Lillicrap \emph{et al.}, ``Continuous control with deep reinforcement learning,'' \emph{arXiv preprint}, arXiv:1509.02971, Sep. 2015. [Online]. Available: \url{https://arxiv.org/abs/1509.02971}.


\bibitem{theile2024equivariant}
M.~Theile, H.~Cao, M.~Caccamo, and A.~L.~Sangiovanni-Vincentelli, 
``Equivariant ensembles and regularization for reinforcement learning in map-based path planning,'' 
in \emph{Proc. IEEE/RSJ Int. Conf. Intell. Robots Syst. (IROS)}, Oct. 2024, pp.~14164--14171.

\bibitem{wang2023ensuring}
H.~Wang, C.~H.~Liu, H.~Yang, G.~Wang, and K.~K.~Leung, 
``Ensuring threshold AoI for UAV-assisted mobile crowdsensing by multi-agent deep reinforcement learning with transformer,'' 
\emph{IEEE/ACM Trans. Netw.}, vol.~32, no.~1, pp.~566--581, 2023.

\bibitem{achiam2017cpo}
J.~Achiam, D.~Held, A.~Tamar, and P.~Abbeel, 
“Constrained policy optimization,” 
in \emph{Proc. Int. Conf. Mach. Learn. (ICML)}, 
Sydney, Australia, Jul. 2017, pp.~22--31.



\end{thebibliography}
\end{document}